\documentclass[letterpaper]{article}
\usepackage{geometry}
\usepackage{amsmath}
\usepackage{amsfonts}
\usepackage{amssymb}
\usepackage{graphicx}
\usepackage{url}
\usepackage{hyperref}
\usepackage{booktabs}
\usepackage{multirow}
\usepackage{makecell}
\usepackage{algorithm}
\usepackage{algpseudocode}
\usepackage{newfloat}
\usepackage{listings}
\usepackage{caption}
\usepackage{natbib}
\usepackage{times}
\usepackage{helvet}
\usepackage{courier}
\usepackage{color}
\usepackage{xcolor}
\usepackage{bm}
\usepackage{dsfont}

\newtheorem{thm}{Theorem}
\newtheorem{proof}{Proof}

\newtheorem{definition}[thm]{Definition}

\renewcommand{\vec}[1]{\bm{#1}}

\newcommand{\Top}{\operatorname{Top}}
\newcommand{\Topg}[3]{\operatorname{Top}_{#1}(#2,#3)}

\DeclareCaptionStyle{ruled}{labelfont=normalfont,labelsep=colon,strut=off}
\floatstyle{ruled}
\newfloat{listing}{tb}{lst}{}
\floatname{listing}{Listing}

\title{Navigating the Exploration–Exploitation Tradeoff in Inference-Time Scaling of Diffusion Models}

\author{Xun Su\textsuperscript{1} \and Jianming Huang \and YANG Yusen \and Zhongxi Fang \and Hiroyuki Kasai}

\date{}

\begin{document}

\maketitle

\setcounter{footnote}{0}
\renewcommand{\thefootnote}{\arabic{footnote}}

\vspace{-1em}
\begin{center}
\textit{Waseda University}
\end{center}
\vspace{0.5em}

\addtocounter{footnote}{1}
\footnotetext{Corresponding author: email: suxun\_opt@asagi.waseda.jp}

\addtocounter{footnote}{1}
\footnotetext{Preprint. Under review.}

\begin{abstract}
    Inference-time scaling has achieved remarkable success in language models, yet its adaptation to diffusion models remains underexplored. We observe that the efficacy of recent Sequential Monte Carlo (SMC)-based methods largely stems from globally fitting the The reward-tilted distribution, which inherently preserves diversity during multi-modal search. However, current applications of SMC to diffusion models face a fundamental dilemma: early-stage noise samples offer high potential for improvement but are difficult to evaluate accurately, whereas late-stage samples can be reliably assessed but are largely irreversible. To address this exploration–exploitation trade-off, we approach the problem from the perspective of the search algorithm and propose two strategies: \textit{Funnel Schedule} and \textit{Adaptive Temperature}. These simple yet effective methods are tailored to the unique generation dynamics and phase-transition behavior of diffusion models. By progressively reducing the number of maintained particles and down-weighting the influence of early-stage rewards, our methods significantly enhance sample quality without increasing the total number of Noise Function Evaluations. Experimental results on multiple benchmarks and state-of-the-art text-to-image diffusion models demonstrate that our approach outperforms previous baselines.
\end{abstract}

\section{Introduction}
Recent years have witnessed significant advancements in generative models across language, image, video, audio, and biological data, driven by scaling laws related to model size, data volume, and computational resources \cite{kaplan2020scalinglawsneurallanguage}. The scaling laws have now extended to the inference-time stage for Large Language Models (LLMs), enhancing performance through variable inference-time computation strategies such as \textit{best-of-N}, \textit{majority voting}, and \textit{tree of thoughts} \cite{qiu2024treebon, zhao2023large, hao2023reasoning, wangself, NEURIPS2023_271db992, loula2025syntactic, wu2025inference}. Such methods have substantially raised the performance ceiling of LLMs in complex reasoning tasks. In contrast, Diffusion Models (DMs), the dominant architecture in image and video generation \cite{pmlr-v37-sohl-dickstein15, ho2020denoisingdiffusionprobabilisticmodels, song2021scorebasedgenerativemodelingstochastic}, have advanced in conditional generation tasks such as text-to-image/voice/video/3D and even diffusion-based LLMs \cite{rombach2022highresolutionimagesynthesislatent, podell2024sdxl, chen2024pixartsigmaweaktostrongtrainingdiffusion, jeong2021diff, chen2024f5,ho2022video, poole2022dreamfusion, nie2025largelanguagediffusionmodels, gat2024discrete}. However, DMs have received relatively little attention with respect to inference-time scaling. They continue to face challenges in complex conditional generation tasks \cite{lu2024handrefiner}, particularly when handling intricate prompts in text-to-image generation, often requiring users to generate multiple outputs to obtain a satisfactory result.

\begin{figure*}[t]
    \centering
    \includegraphics[width=\textwidth]{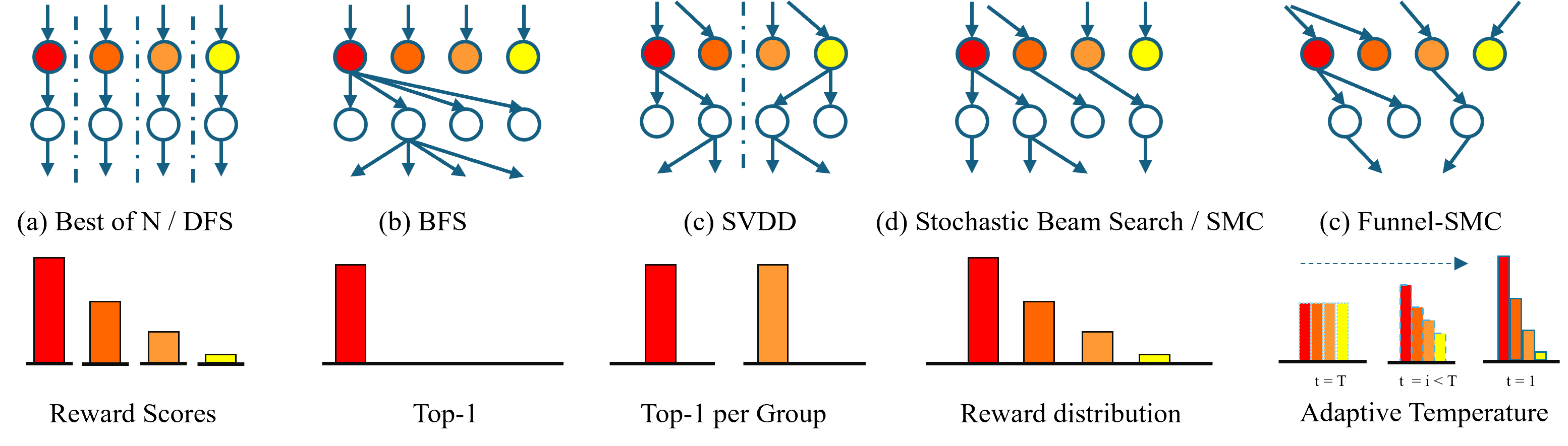}
    \caption{Illustration of inference-time scaling methods. The intensity of node color indicates the selection weight, with darker colors representing higher weights. \textit{F-SMC} progressively reduces the number of particles during the process, while \textit{Adaptive Temperature} smooths the weight distribution in early stages by lowering the temperature parameter \(\lambda_t\) in the potential function (see Eq.~\ref{eq:target_distribution}), thereby reducing the impact of inaccurate early-stage rewards.}

    \label{fig:fsmc}
\end{figure*}

\begin{figure}[t]
    \centering
    \includegraphics[width=1.0\linewidth]{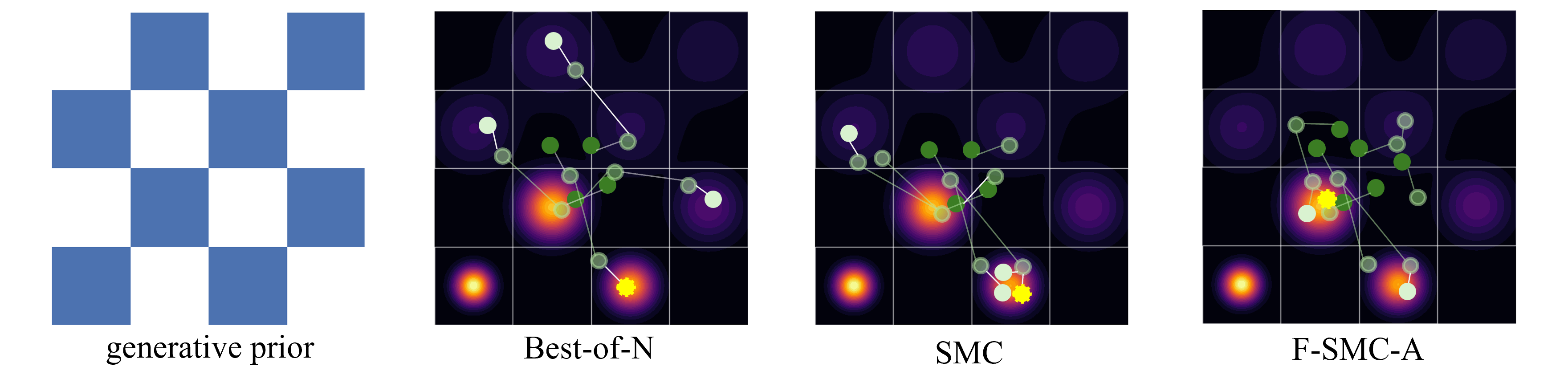}
    \caption{Illustration of the proposed method. The green color indicates the starting point, and the color gradually lightens as the sampling process of the DMs progresses with decreasing $t$. The yellow point with jagged edges represents the final selected best sample.}
    \label{fig:method}
\end{figure}

Considering that a reward model can evaluate the quality of generated images, inference-time scaling aims to produce images with higher scores within a fixed computational budget. Unlike Reinforcement Learning (RL) methods, which train the generative model to align with the reward distribution \cite{wallace2024diffusion, liu2025flowgrpotrainingflowmatching, fan2023dpok}, the inference-time scaling in DMs can be viewed as a search problem in a multi-modal, high-dimensional space. The solution for DMs depends on the selection of noise vectors at each Stochastic Differential Equation (SDE) solvers step throughout the generation process. The well-known \textit{best-of-N} approach can be regarded as a Depth-First Search (DFS) method~\cite{tarjan1972dfs}, whereas \textit{Soft Value-Based Decoding} (SVDD)~\cite{li2024derivativefreeguidancecontinuousdiscrete} resembles a Breadth-First Search (BFS) within a given noise group, selecting the best noise from multiple candidates at each iteration~\cite{moore1959shortest}. Recent works~\cite{singhal2025generalframeworkinferencetimescaling, li2024derivativefreeguidancecontinuousdiscrete} have framed inference-time scaling of DMs within the Feynman–Kac formalism, proposing the use of Sequential Monte Carlo (SMC) methods to approximate the reward-tilted distribution. This formulation underpins many subsequent approaches.

In this paper, we interpret SMC-based methods as a form of stochastic Beam Search, which preserves diversity during the search by leveraging sampling weights that theoretically guide the generative distribution toward the target distribution. We highlight an inherent exploration–exploitation trade-off in inference-time scaling for DMs, a challenge largely overlooked in prior work:

\begin{quote}
\textit{Early-stage noises are hard to evaluate but still tiltable, while late-stage noises are easier to score but harder to tilt.}
\end{quote}

\begin{figure}[t]
    \centering
    \begin{minipage}{0.48\textwidth}
        \centering
        \includegraphics[width=\textwidth]{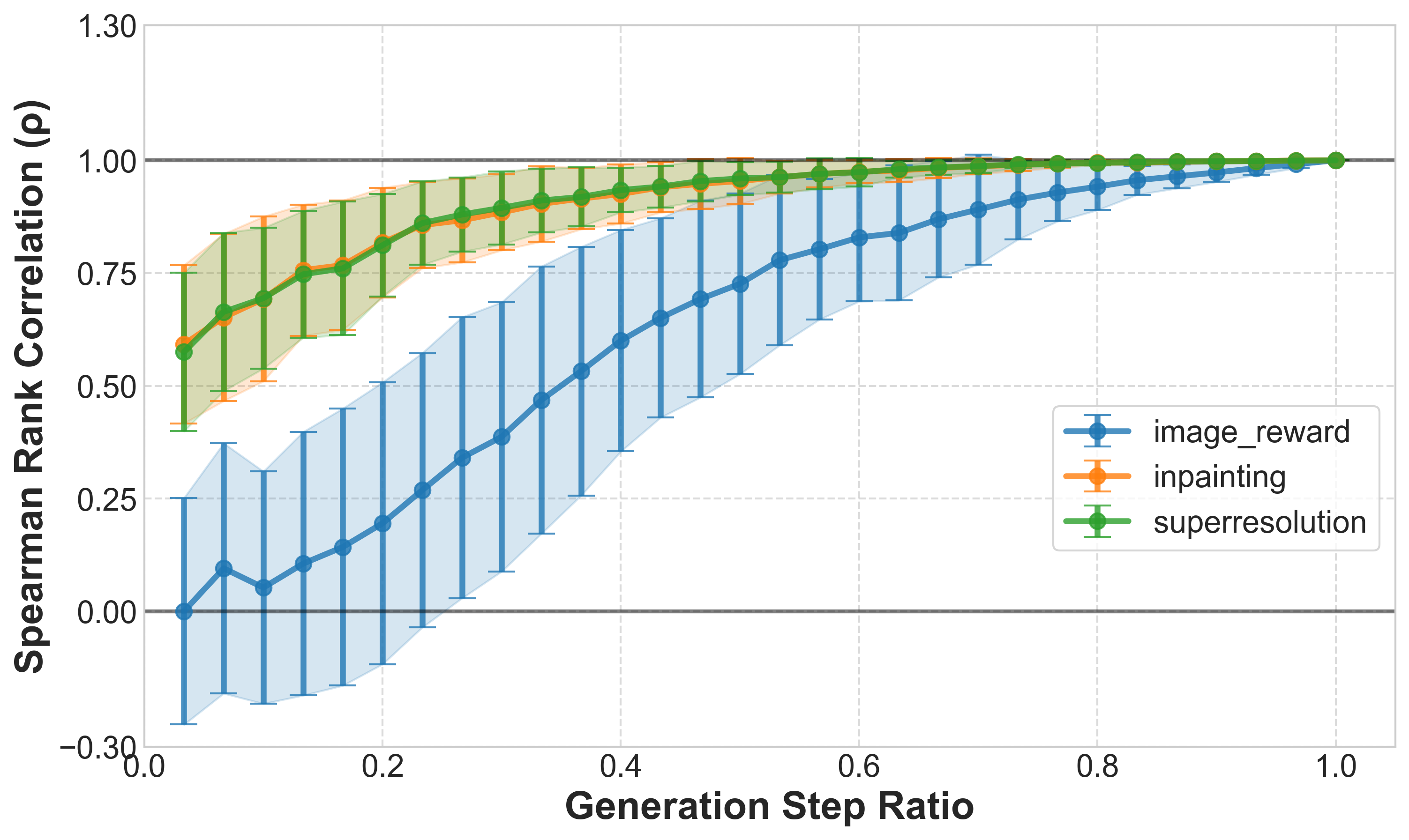}
        \caption{Rank correlation of verifier scores between intermediate timesteps and final outputs across different reward functions.}
        \label{fig:rank_correlation}
    \end{minipage}
    \hfill
    \begin{minipage}{0.48\textwidth}
        \centering
        \includegraphics[width=\textwidth]{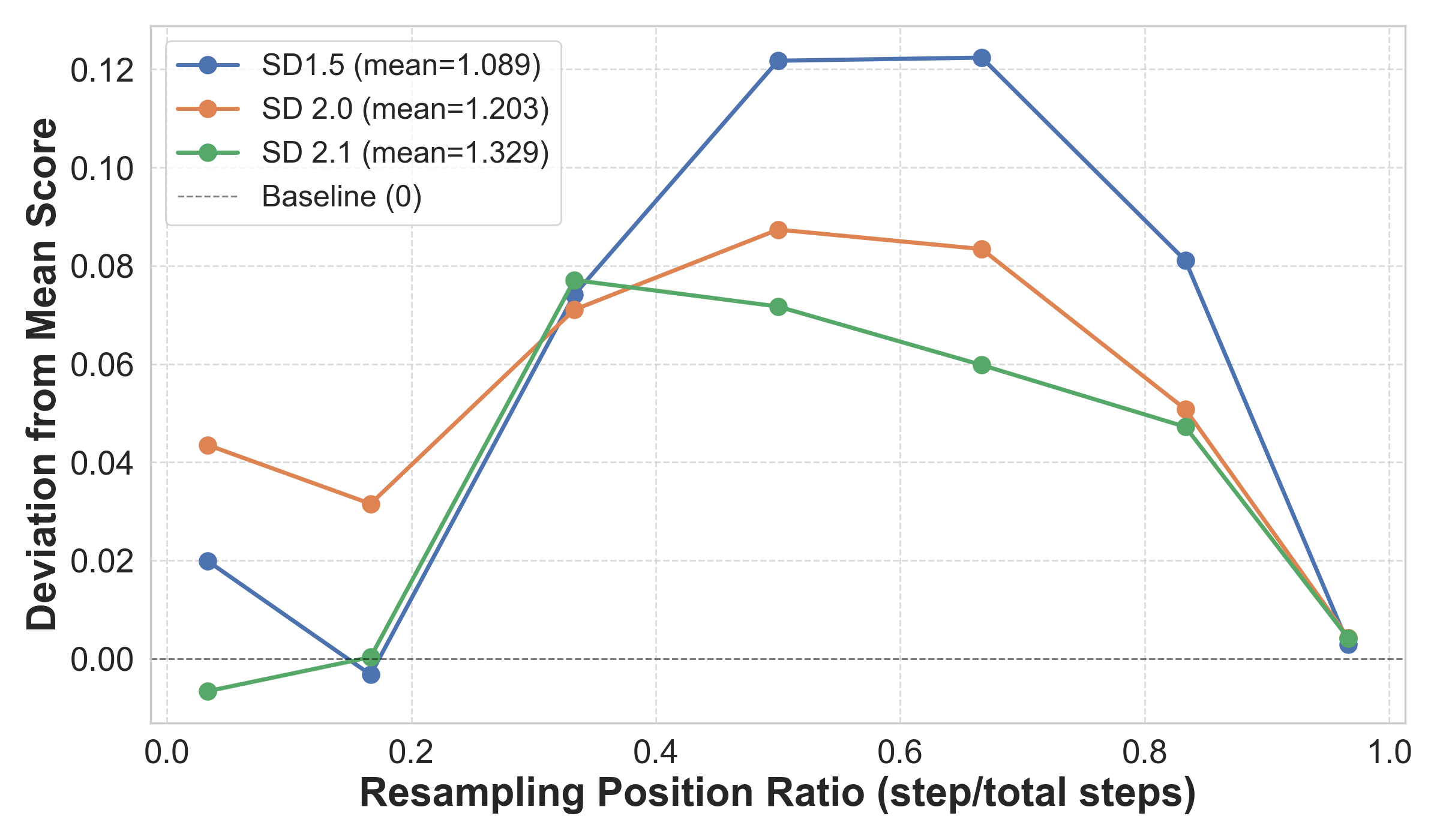}
        \caption{Effect of resampling timing on image quality.}
        \label{fig:influence}
    \end{minipage}
\end{figure}

The expected outputs of early-stage particles are tend to be blurry. Since reward models are typically trained on high-quality, sharp images, they struggle to accurately assess the quality of these blurry intermediate samples. In the context of text-to-image alignment (Figure~\ref{fig:rank_correlation}), scores from reward model are significantly less reliable in early stages compared to those from general inverse-problem verifiers. As shown in Figure~\ref{fig:influence}, the reward scores during sampling forms a concave curve. Our key insight is that current SMC-based methods are limited by a structural dilemma in diffusion generation: (1) reward model cannot reliably evaluate early-stage images despite their long-term importance; and (2) once the image structure solidifies in later stages, resampling interventions become largely ineffective.

Motivated by this observation, based on the success of SMC-based methods \cite{singhal2025generalframeworkinferencetimescaling}, we propose two methods: \textbf{\textcolor{blue}{Funnel-SMC (F-SMC)}} and an \textbf{\textcolor{red}{Adaptive Temperature (SMC-A)}} parameter strategy.

\textbf{\textcolor{blue}{F-SMC}} adopts a funnel-shaped particle count schedule that decreases over the sampling process. It allocates more particles for exploration during the early-to-middle stages and gradually reduces the count as the image structure becomes fixed,  to efficiently reallocate computational resources in the early stages while reducing cost in later stages.

\textbf{\textcolor{red}{SMC-A}} addresses the issue of inaccurate early-stage reward estimation by introducing a dynamic temperature parameter $\lambda_t$ into the potential function design of SMC. By adaptively increasing $\lambda_t$ as the sampling progresses, it mitigates the degeneration of particle diversity and enhances reward scores throughout the generation process.

Our contributions are summarized as:

\begin{itemize}

    \item We identify the fundamental exploration–exploitation trade-off in DM generation: early-stage samples are highly malleable but difficult to evaluate, while late-stage samples are easier to assess but resistant to meaningful modification. Through extensive experiments and toy model analysis, we demonstrate how this trade-off exists and inherently limits the effectiveness of current inference-time scaling methods.

    \item We propose two enhancements to standard SMC approaches: \textbf{\textcolor{blue}{F-SMC}}, which adapts the particle count to match the \textcolor{blue}{decreasing plasticity} of the generation process; and \textbf{\textcolor{red}{SMC-A}}, which incorporates a dynamic temperature parameter into the potential function design to counter \textcolor{red}{early estimation inaccuracies}. We further provide a theoretical analysis proving the convergence of our method, enabling principled and effective application of SMC to inference-time scaling in DMs.

    \item Extensive experiments across multiple models and evaluation metrics demonstrate that our approach consistently outperforms both classic search strategies and prior SMC-based methods~\cite{singhal2025generalframeworkinferencetimescaling}. Notably, simply using standard SMC with the common \textit{Diff} weight achieves state-of-the-art performance within our framework.

\end{itemize}

\begin{figure}[t]
    \centering
    \hspace{0.2cm} Single \hspace{1.2cm} Best-of-N \hspace{1cm} Top-$k$ \hspace{1.3cm} SVDD \hspace{1.2cm} SMC(M) \hspace{0.7cm} F-SMC(M) \hspace{0.4cm} F-SMC-A(D)\\
    \includegraphics[width=1.0\textwidth]{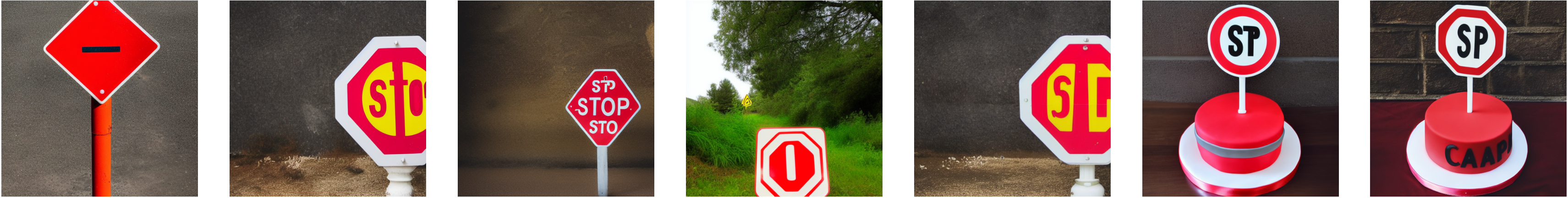}\\
    \textcolor{red}{a photo of a cake and a stop sign} 
    \includegraphics[width=1.0\textwidth]{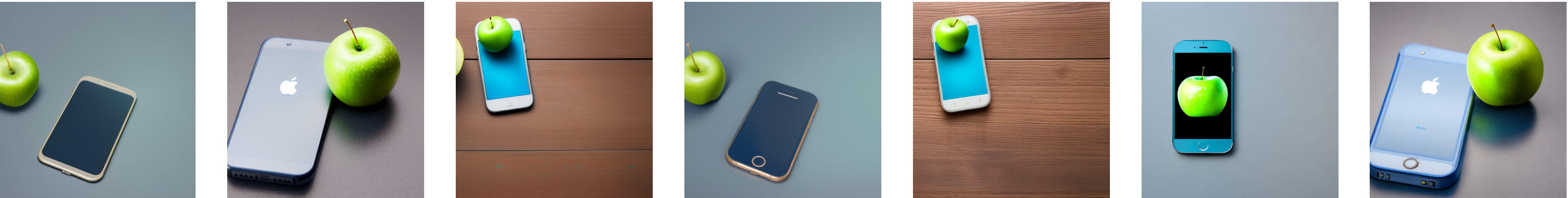}\\
    \textcolor{blue}{a photo of a blue cell phone and a green apple} 
    \includegraphics[width=1.0\textwidth]{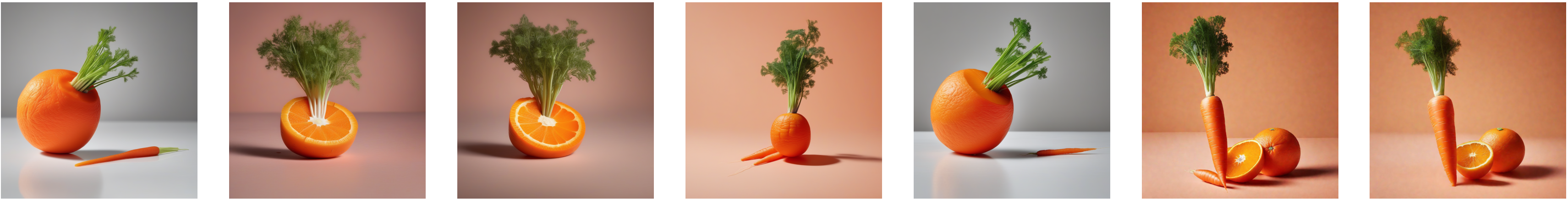}\\
    \textcolor{orange}{a photo of a carrot left of an orange}
    \caption{Generated images using different methods: (a) Single; (b) Best-of-N; (c) Top-$k$; (d) SVDD; (e) SMC (M); (f) F-SMC (M); and (g) F-SMC-A (D). We select three prompts from the Geneval dataset, each corresponding to a distinct challenging tag: \textit{\textcolor{red}{a photo of a cake and a stop sign}} from the \textbf{\textcolor{red}{two objects}} category, generated using \textcolor{red}{SD 1.5}; \textit{\textcolor{blue}{a photo of a blue cell phone and a green apple}} from the \textbf{\textcolor{blue}{color attribution}} category, generated using \textcolor{blue}{SD 2.1}; and \textit{\textcolor{orange}{a photo of a carrot left of an orange}} from the \textbf{\textcolor{orange}{position}} category, generated using \textcolor{orange}{SDXL}.}
    \label{fig:generated_images}
\end{figure}

\section{Related Work}

\paragraph{Noise Searching and Inference-Time Scaling}

The randomness in DMs generation arises from the stochastic nature of SDE-based samplers. When using an Ordinary Differential Equation (ODE) solver instead, the generation process becomes deterministic—each output corresponds one-to-one with its initial noise vector. This leads to two primary research directions in inference-time scaling:

\begin{enumerate}
    \item \textbf{ODE-based methods} aim to discover favorable initial noise, often termed "golden noise", that are more likely to produce high-quality, prompt-consistent outputs. These methods include first and zeroth-order optimization techniques~\cite{chefer2023attend, guo2024initno, chen2024tino, bai2024zigzagdiffusionsamplingdiffusion}, as well as DDIM inversion approaches~\cite{bai2024zigzagdiffusionsamplingdiffusion} that attempt to trace back from target concepts to initial noise. However, these approaches suffer from instability, weak theoretical foundations~\cite{ma2025inference}, and significant computational overhead compared to simple \textit{best-of-N} sampling~\cite{bai2024zigzagdiffusionsamplingdiffusion, qi2024noisescreatedequallydiffusionnoise}.

    \item \textbf{SDE-based methods} treat inference-time scaling as a stochastic search process over sampling trajectories. Initiated by~\cite{ma2025inference}, this line of work explores techniques such as derivative-free guidance, stochastic control, and SMC to navigate the space of possible noise trajectories during generation. These methods have been extended to large-scale diffusion and flow-matching models for both image and video synthesis~\cite{xie2025sana15efficientscaling, li2024derivativefreeguidancecontinuousdiscrete, kim2025inferencetimescalingflowmodels}.
\end{enumerate}

\paragraph{SMC-based Sampling and Gradient Guidance}

SMC methods were first introduced into DMs in the context of solving inverse problems, where conditioning information is used to guide the sampling process. Unlike traditional gradient-based guidance methods~\cite{chung2023diffusion, wang2022zero, wang2023zeroshot, mardani2023variational, song2023pseudoinverseguided, mardani2024a, yu2023freedom, kim2025regularization}, SMC-based approaches bypass direct gradient computation. Instead, they consider measurement consistency as weight and resample. This strategy preserves image coherence while avoiding instability often associated with gradient injection. The use of SMC in diffusion-based inverse problem solving was first demonstrated by~\cite{wu2023practical, dou2024diffusion}, and later extended by~\cite{singhal2025generalframeworkinferencetimescaling} to more general guided generation tasks, now commonly referred to as inference-time scaling.

\section{Background}

\subsection{Diffusion Models and SDE-based Samplers}
DMs define a forward process $q(\vec{x}_t|\vec{x}_{t-1})$ that incrementally adds Gaussian noise to clean data $\vec{x}_0$:
\begin{equation}
 q(\vec{x}_t|\vec{x}_{t-1}) = \mathcal{N}(\vec{x}_t; \sqrt{1-\beta_t}\vec{x}_{t-1}, \beta_t I),
\end{equation}
where $\beta_t$ denotes the noise schedule. This forward process transforms clean samples $\vec{x}_0$ into pure noise $\vec{x}_T$ over $T$ steps. The generative process, by contrast, is a reverse-time Markov chain $\vec{x}_T \rightarrow \vec{x}_{T-1} \rightarrow \cdots \rightarrow \vec{x}_0$ that gradually removes noise and recovers the data distribution~\cite{song2021scorebasedgenerativemodelingstochastic, ho2020denoisingdiffusionprobabilisticmodels}. The marginal likelihood is given by:
\begin{equation}
 p(\vec{x}_0) = \int \prod_{t=1}^{T} p(\vec{x}_{t-1} | \vec{x}_t) p(\vec{x}_T) \, \mathrm{d}\vec{x}_T \cdots \mathrm{d}\vec{x}_1.
\end{equation}

A neural network $\theta$ is trained to approximate the reverse transition $p_\theta(\vec{x}_{t-1} \mid \vec{x}_t)$ using a score-based formulation:
\begin{equation}
 p_{\theta}(\vec{x}_{t-1} | \vec{x}_t) = \mathcal{N}(\vec{x}_{t-1}; \vec{x}_t + \sigma_t s_\theta(\vec{x}_t, t), \sigma_t^2 I),
\end{equation}
where $\sigma_t$ is the noise scale at timestep $t$, and $s_\theta(\vec{x}_t, t)$ approximates the score function $\nabla_{\vec{x}_t} \log p(\vec{x}_t)$.

It is important to emphasize that once the timestep discretization is fixed, generation using an SDE-based sampler is fully specified by an initial noise vector $\vec{x}_T$ and a sequence of noise terms $\epsilon_T, \ldots, \epsilon_1$. We refer to the selection of this noise sequence as the \textbf{generation strategy} $\pi$.


\section{Funnel-SMC Method and Adaptive Temperature}

\subsection{Inference-Time Scaling as Sequential Search}\label{sec:seq-search}

At inference, we are given a prompt $\mathbf{c}$ and a (typically imperfect) reward model $r(\cdot, \mathbf{c})$. Let $\mathcal{E}$ denote the set of all generation strategies that respect a pre-specified compute budget (e.g., NFE). The objective of \emph{inference-time scaling} is to discover an optimal strategy $\pi^{\star}$ that maximizes an unknown ground-truth reward $r^{\star}$:
\begin{equation}
r^{\star}(\vec{x}(\pi^{\star}), \mathbf{c}) = \max_{\pi\in\mathcal{E}}r(\vec{x}(\pi), \mathbf{c}).\label{eq:objective}
\end{equation}

where we use $\vec{x}(\pi)$ to denote the image generated by the strategy $\pi$. Because every choice of noise (and its subsequent refinement) can be viewed as a node in a search tree of depth~$T$. At each layer $t$ the algorithm maintains a population $\{(\vec{x}_t^{(i)},w_t^{(i)})\}_{i=1}^{N_t}$, $N_t$ is the \emph{population size}. Here, we introduce a framework to unify the search procedure of inference-time scaling on DMs.

\paragraph{General Search Framework of Inference-Time Scaling} 
\label{sec:general_search_framework}
The inference-time scaling procedure can be abstracted as a sequential search process composed of three key operators:

\begin{enumerate}
    \item \textbf{Selection} (\emph{``resample''}): At step $t$, given $N_t$ particles (with $N_{t-1} < N_t$), indexed from $1$ to $N_t$, we sample a new index set $\mathcal{I}_{t-1} \in \{1, \dots, N_t\}^{N_{t-1}}$ according to a resampling policy $\mathcal{R}$ based on normalized weights $w_t$:
    \begin{equation}
    \{\vec{x}_t^{(i)}\}_{i = 1}^{N_{t-1}} \sim \mathcal{R}\big(\{\vec{x}_t^{(j)}, w_t^{(j)}\}_{j=1}^{N_t}\big).
    \end{equation}

    \item \textbf{Transition} (\emph{``propagate''}): For each selected particle, a successor is sampled from a proposal distribution:
    \begin{equation}
    \vec{x}_{t-1}^{(i)} \sim \tau(\vec{x}_{t-1} \mid \vec{x}_t^{(i)}),
    \end{equation}
    where $\tau$ is the proposal kernel that propagates particles from timestep $t$ to $t-1$. The time-reversed diffusion kernel is a standard choice, though guided or twisted proposals may also be used.

    \item \textbf{Scoring} (\emph{``weight''}): Each resulting particle is assigned a non-negative importance weight:
    \begin{equation}
    w_{t-1}^{(i)} \propto G_{t-1}\bigl(\vec{x}_{\rho_{t-1}(i, T:t-1)}, \mathbf{c}, \lambda \bigr),
    \end{equation}
    Here, $G_{t-1}$ is a \emph{potential function} that evaluates the trajectory history of particle $\vec{x}_{t-1}^{(i)}$. The function $\rho_t(i, s)$ denotes the ancestor index at timestep $s$ of the $i$-th particle at time $t$, with $\rho_t(i, t) = i$. The notation $\vec{x}_{\rho_t(i, T:t-1)}$ represents the full ancestral path from timestep $T$ down to $t-1$ leading to the current particle.
\end{enumerate}

\begin{table}[t]
    \centering
    \caption{Selection rules and potential function definitions of different methods under the unified search framework. $\Top(\cdot, K)$ selects the top $K$ particles by weight. $\Topg{g}{\cdot}{K}$ groups particles by $g$ and selects the top $K$ within each group, followed by replication to match the original group size. We use $r_{t}$ to denote the reward function $r$ computed based on $\vec{x}_{0|t}$ (see Def~\ref{def:tweedie}). \textbf{SMC(D)} and \textbf{SMC(M)} denotes the use of \emph{Diff} and \emph{Max}, respectively.}
    \setlength{\tabcolsep}{2pt}
    \begin{tabular}{lcc}
      \toprule
      Method & Selection $\mathcal{R}_t$ & Potential $G_t$ \\
      \midrule
  
      \textbf{DFS} &
        $\mathcal{I}_t = \mathcal{I}_{t-1}$ &
        $r_t$ \\[4pt]
  
      \textbf{BFS} &
        $\mathcal{I}_t = \Top(\cdot, K)$ &
        $r_t$ \\[4pt]
  
      \textbf{SVDD} &
        $\displaystyle
        \mathcal{I}_t = \bigcup_g \Topg{g}{\cdot}{K}$ &
        $r_t$ \\[4pt]
  
      \textbf{SMC(D)} &
        $\Pr(i \in \mathcal{I}_t) = w_t^{(i)}$ &
        $\exp[\lambda (r_t - r_{t-1})]^{\dag}$ \\[2pt]
  
      \textbf{SMC(M)} & same as above &
        $\exp[\lambda \max_{T > j \geq t} r_j]^{\ddag}$ \\[2pt]
  
      \bottomrule
    \end{tabular}\\
    {\raggedright 
    $^{\dag  \ddag}$ When $t = 0$ or $t = T$, the potential functions are defined with special cases; see the Theorem~\ref{thm:converge} for details.\par}
    \label{tab:sel-pot}
  \end{table}
In the context of inference-time scaling for DMs, we exclude backtracking-style algorithms from consideration \cite{bai2024zigzagdiffusionsamplingdiffusion, zhang2025inferencetimescalingdiffusionmodels, ma2025inference}, as they are not suited for parallel execution, introduce additional procedural complexity, and might benefits implicitly from increasing classifier-free guidance (CFG) scale. \cite{bai2024zigzagdiffusionsamplingdiffusion}.

We aim to interpret inference-time scaling algorithms within the unified three-stage framework described above: \emph{selection}, \emph{transition}, and \emph{scoring}. For the transition stage, we focus exclusively on the native reverse process learned by the pretrained DMs, $p_\theta(\vec{x}_{t-1} | \vec{x}_t)$, and do not incorporate guided transitions. It is motivated by practical consideration: gradient computations scale poorly with model size and become increasingly slow and unstable in large-scale models. By adhering to the original denoising dynamics, we ensure consistency across different algorithmic comparisons.

The diversity of inference-time scaling strategies can be expressed through variations in the selection policy $\mathcal{R}$ and the potential function $G_t$, as summarized in Table~\ref{tab:sel-pot}. The goal of SMC-based methods is to steer the particle system toward a target distribution $p_{\text{tar}}$:
\begin{equation}
\label{eq:target_distribution}
p_{\text{tar}} = \frac{1}{Z}p_{\text{pre}}(\vec{x}, \mathbf{c}) \cdot \exp(\lambda r(\vec{x}, \mathbf{c})),
\end{equation}
where $p_{\text{pre}}$ denotes the generative prior, $\lambda$ is the temperature and $r(\cdot, \mathbf{c})$ is the reward model conditioned on prompt $\mathbf{c}$.

SMC-based methods share a common probabilistic resampling mechanism, where particles are selected in proportion to their weights. Methods diverge in how they define the potential function $G_t$ used to compute those weights. Two widely used forms—the \emph{difference potential} (Diff) and the \emph{maximum potential} (Max)—are summarized in Table~\ref{tab:sel-pot}.

The canonical SMC formulation adopts the \emph{Diff}, which ensures that particle weights reflect local reward increments. In contrast, the \emph{Max}, has been shown to outperform alternative strategies empirically \cite{singhal2025generalframeworkinferencetimescaling}. By emphasizing the max reward achieved along a particle’s ancestor trajectory, it retains high-reward candidates all the time, relying on a final-step renormalization to maintain consistency with the target distribution.

The success of SMC in DMs largely stems from its ability to preserve sample diversity while guiding exploration toward high-reward regions. This is particularly important in the presence of phase transition phenomena observed in diffusion sampling trajectories~\cite{pavasovic2025understandingclassifierfreeguidancehighdimensional, li2024criticalwindows}, where semantic commitment to specific concepts or modes occurs abruptly. Maintaining diversity prior to this transition is critical, as it determines whether the generative path will converge to a desirable mode under a limited sampling budget.

Building on this insight, our proposed F-SMC allocates more particles to early stages—where the probability of discovering promising trajectories is highest—and progressively reduces the population as the image structure stabilizes. Notably, we find that the \textbf{Diff} performs poorly in practice due to inaccurate early-stage reward estimates, which are further exaggerated by the difference operation. To address this, our proposed \textbf{SMC-A} employs an adaptive temperature to mitigate the impact of unreliable early scores. We validate the effectiveness of both F-SMC and SMC-A through direct comparisons in Tables~\ref{tab:smc_comparison1} and~\ref{tab:smc_comparison2}.

\subsection{Proposed Algorithm Overview}
\textbf{F-SMC} introduces a dynamic particle count schedule $\{N_t\}_{t=0}^T$, where the particle count $N_t$ decreases monotonically over time ($N_{t-1} < N_t$). The \textbf{SMC-A} defines a time-dependent reward scaling factor $\lambda_t = \frac{T - t}{T} \lambda_0$, which linearly increases the influence of the reward from 0 to the defined temperature in $p_{\text{tar}}$. We use \textbf{F-SMC-A} to represent a combination of two proposals.

To improve reward estimation, we apply the Tweedie formula to obtain a denoised prediction $\vec{x}_{0|t}$ from $\vec{x}_t$, and use $\vec{x}_{0|t}$ as input to the $r(\cdot | \vec c)$ in all methods \cite{dieleman2024spectral, kadkhodaie2021stochastic}. The full \textbf{F-SMC-A} algorithm is described in detail in Algorithm~\ref{alg:fsmc}.

\begin{definition}[Tweedie Denoising Estimate]
    \label{def:tweedie}
    Let $\vec{x}_t$ be the noisy sample at timestep $t$.
    Define the cumulative noise schedule $\bar{\alpha}(t)=\prod_{j=1}^{t}(1-\beta_j)$. The Tweedie formula gives the following approximation of the $\mathbb{E}\!\left[\vec{x}_0 | \vec{x}_t\right]$:
    \begin{equation}
    {\vec{x}}_{0\mid t}
    \;=\;
    \frac{1}{\sqrt{\bar{\alpha}(t)}}
    \Bigl(
    \vec{x}_t
    +
    \bigl(1-\bar{\alpha}(t)\bigr)\,
    s_\theta(\vec{x}_t,t)
    \Bigr).
    \end{equation}
\end{definition}

\begin{algorithm}[t]
    \caption{F-SMC-A Diffusion Sampling}
    \label{alg:fsmc}
    \begin{algorithmic}[1]
    \Require Text prompt $\vec c$, DM $\vec s_\theta$, verifier $r$, target temperature $\lambda$
    \Require Guidance scale $w$, particle schedule $\{N_t\}_{t=0}^T$ with $N_T > \cdots > N_0$, number of sampling steps $T$
    \Ensure Optimal generated image $\vec{x}_{\text{best}}$
    
    \State Sample initial noise vectors $\{\vec{x}_T^{(i)}\}_{i=1}^{N_T} \sim \mathcal{N}(0, \mathbf{I})$
    \State Initialize $\lambda_T \gets 0$
    \State Compute initial scores $w_T^{(i)} \gets G_T(\vec{x}_T^{(i)}, \vec c, \lambda_T)$
    \For{$t \gets T$ \textbf{down to} $1$}
        \State Set $\lambda_t \gets \frac{T - t}{T} \lambda$ \Comment{Adaptive temperature}
        \State Resample $\{\vec{x}_t^{(i)}\}_{i=1}^{N_{t-1}} \gets \mathcal{R}(\{\vec{x}_t^{(j)}, w_t^{(j)}\}_{j=1}^{N_t})$
        \For{$i \gets 1$ \textbf{to} $N_{t-1}$}
            \State Propagate $\vec{x}_{t-1}^{(i)}, \vec{x}_{0|t}^{(i)} \gets \tau(\vec{x}_{t-1} \mid \vec{x}_t^{(i)},\vec c)$
            \State Weight $w_{t-1}^{(i)} \gets r(\vec{x}_{0|t}^{(i)}, \vec c)$
        \EndFor
    \EndFor
    
    \State Compute final scores $s_i \gets r(\vec{x}_0^{(i)}, \vec c)$ for $i = 1, \dots, N_0$
    \State $\text{best idx} \gets \arg\max_i^{i<N_0} s_i$
    \State \Return $\vec{x}_{\text{best}} \gets \vec{x}_0^{\text{best idx}}$
    \end{algorithmic}
    \end{algorithm}

\subsection{Convergence Analysis}

\begin{thm}[Convergence of F-SMC-A]
\label{thm:converge}
Let $p_{\theta}(\vec{x}_{t:T}, \vec c)$ denote the conditional trajectory distribution of a pretrained DM, $\lambda$ as the temperature of reward probability, and let $G_t$ be a non-negative potential function such that
\[
\prod_{t=T}^{0} G_t(\vec{x}_{T:t}, \vec c) = \exp(\lambda r(\vec{x}_0, \vec c)),
\]
for some bounded reward function $r$. Then, as the number of particles $N_t \to \infty$ at each timestep $t$, the empirical distribution defined by the F-SMC-A algorithm converges in distribution to the target:
\begin{equation}
\label{eq:target_distribution_theorem}
p_{\mathrm{tar}}(\vec{x}_{0}, \vec c) \propto p_{\mathrm{pre}}(\vec{x}_{0}, \vec c) \exp(\lambda r(\vec{x}_0, \vec c)).
\end{equation}
\end{thm}

\begin{table*}[t]
    \centering
    \caption{Performance comparison across different models and Inference-time scaling methods.}
    \label{tab:inference_scaling}
    \begin{tabular}{lcccccccc}
    \toprule
    \textbf{Model} & \textbf{Metric} & \textbf{Best-of-N} & \textbf{BFS} & \textbf{SVDD} & \textbf{SMC(M)} & \textbf{SMC(D)} & \textbf{F-SMC-A(M)} & \textbf{F-SMC-A(D)} \\
    \midrule
    \multirow{4}{*}{SD 1.5} & IR $\uparrow$ & 1.003 & 1.007 & 0.9152 & 1.000 & 0.9590 & 1.007 & \textbf{1.0785} \\
    & HPSv2 $\uparrow$ & \textbf{0.2899} & 0.2867 & 0.2875 & 0.2878 & 0.2862 & {0.2883} & 0.2891 \\
    & \textcolor{blue}{GenEval} $\uparrow$ & \textcolor{blue}{56.24} & \textcolor{blue}{57.14} & \textcolor{blue}{54.25} & \textcolor{blue}{55.52} & \textcolor{blue}{54.97} & \textcolor{blue}{56.78} & \textcolor{blue}{\textbf{57.32}} \\
    \midrule
    \multirow{4}{*}{SD 2.0} & IR $\uparrow$ & 1.174 & 1.187 & 1.150 & 1.197 & 1.158 & 1.249 & \textbf{1.258} \\
    & HPSv2 $\uparrow$ & \textbf{0.2973} & 0.2947 & 0.2949 & 0.2959 & 0.2940 & 0.2950 & {0.2966} \\
    & \textcolor{blue}{GenEval} $\uparrow$ & \textcolor{blue}{59.86} & \textcolor{blue}{60.76} & \textcolor{blue}{59.86} & \textcolor{blue}{61.12} & \textcolor{blue}{56.24} & \textcolor{blue}{{60.76}} & \textcolor{blue}{\textbf{62.39}} \\
    \midrule
    \multirow{4}{*}{SD 2.1} & IR $\uparrow$ & 1.241 & 1.277 & 1.198 & 1.2340 & 1.222 & 1.228 & \textbf{1.308} \\
    & HPSv2 $\uparrow$ & 0.2977 & 0.2957 & 0.2960 & 0.2962 & 0.2956 & 0.2958 & \textbf{0.2981} \\
    & \textcolor{blue}{GenEval} $\uparrow$ & \textcolor{blue}{60.04} & \textcolor{blue}{59.67} & \textcolor{blue}{61.84} & \textcolor{blue}{60.04} & \textcolor{blue}{60.94} & \textcolor{blue}{61.48} & \textcolor{blue}{\textbf{61.84}} \\
    \midrule
    \multirow{4}{*}{SDXL} & IR $\uparrow$ & 1.338 & 1.417 & 1.362 & 1.365 & 1.369 & 1.343 & \textbf{1.430} \\
    & HPSv2 $\uparrow$ & \textbf{0.3132} & 0.3089 & 0.3114 & 0.3111 & 0.3098 & 0.3084 & 0.3112 \\
    & \textcolor{blue}{GenEval} $\uparrow$ & \textcolor{blue}{64.01 } & \textcolor{blue}{64.56} & \textcolor{blue}{64.92} & \textcolor{blue}{63.83} & \textcolor{blue}{63.65} & \textcolor{blue}{65.64} & \textcolor{blue}{\textbf{66.55}} \\
    \bottomrule
    \end{tabular}
\end{table*}

\begin{proof}
    \label{proof: potential}
    We follow the convergence theory as outlined in~\cite{singhal2025generalframeworkinferencetimescaling}. The reward-tilted distribution over trajectories from $t$ to $T$ is defined as:
    \begin{equation}
    p_{\mathrm{SMC}, t}(\vec{x}_{T:t}, \vec{c}) = \frac{1}{Z_t} p_{\theta}(\vec{x}_{T:t}, \vec{c}) \prod_{i=T}^{t} G_i(\vec{x}_{T:i}, \vec{c}),
    \end{equation}
    where $Z_t = \mathbb{E}_{p_{\theta}}[\prod_{i=T}^t G_i]$ is the normalization constant. Under suitable regularity conditions (e.g., bounded $G_t$ with finite variance), standard results guarantee that the empirical distribution at $t=0$ converges to the marginal:
    \begin{equation}
    p_{\mathrm{SMC},0}(\vec{x}_{T:0}, \vec{c}) = p_{\mathrm{tar}}(\vec{x}_0, \vec{c}),
    \end{equation}
    as long as the product of potential functions satisfies:
    \begin{equation}
    \label{eq:potential_product}
    \prod_{t=T}^{0} G_t(\vec{x}_{T:t}, \vec{c}) = \exp(\lambda r(\vec{x}_0, \vec{c})).
    \end{equation}
    
    \paragraph{Max}
    For each particle $i \in [1, N_0]$, the potential at $t = 0$ is defined as:
    \[
    G_0^{(i)} = \frac{\exp(\lambda r(\vec{x}_0^{(i)}, \vec{c}))}{\prod_{s=1}^{T} G_{s-1}^{(i)}}.
    \]
    Then the full product of potentials becomes:
    \begin{align}
    \prod_{t=T}^{0} G_t^{(i)} 
    &= G_0^{(i)} \prod_{t=T}^{1} \exp\left( \lambda \max_{\vec{x} \in \vec{x}_{\rho_t(i, T:t)}} r(\vec{x}, \vec{c}) \right) \notag \\
    &= \exp(\lambda r(\vec{x}_0^{(i)}, \vec{c})). \notag
    \end{align}
    \paragraph{Diff}
    For each $i \in [1, N_0]$, the potential at $t = T$ is initialized as:
    $G_T^{(i)} = 1/N_T$. Then,
    \begin{align}
    \prod_{t=T}^{0} G_t^{(i)} 
    &= G_T^{(i)} \prod_{t=T-1}^{0} \frac{\exp(\lambda_t r(\vec{x}_{\rho_0(i, t)}, \vec{c}))}{\exp(\lambda_{t+1} r(\vec{x}_{\rho_0(i, t+1)}, \vec{c}))} \notag \\
    &= \exp(\lambda r(\vec{x}_0^{(i)}, \vec{c})),
    \end{align}
    also satisfying Eq.~\eqref{eq:potential_product}. Thus, the convergence result holds for both Max and Diff potentials.
    
    Hence, F-SMC-A converges $p_{\mathrm{tar}}(\vec{x}_0, \vec{c})$ as $N_0 \to \infty$, thereby establishing the convergence of the F-SMC-A algorithm under the condition that the particle ratio $N_0 / N_t > 0$ for all $t$, for a more in-depth proof check out the appendix.

    We note that the \textit{Diff} potential formulation satisfies the potential product condition in Eq.~\eqref{eq:potential_product} by construction, whereas the \textit{Max} potential requires the inclusion of a normalization factor at $t = 0$ (i.e., $G_0$) to ensure the same theoretical guarantee.

\end{proof}

\section{Experiments}

\paragraph{Experimental Setup}
To evaluate the performance of the proposed methods, we conducted experiments on the text-to-image generation task. 

\textbf{Pre-trained Model}: We utilized pre-trained Stable Diffusion 1.5, 2.0, 2.1, and SDXL \cite{rombach2022highresolutionimagesynthesislatent, podell2024sdxl}. 

\textbf{Sampling Process}: All models were sampled using 30 denoising steps with the Euler Ancestral solver (DDPM) and a classifier-free guidance scale of 7.5.

\textbf{Datasets}: We evaluated on DrawBench~\cite{8803821} and GenEval~\cite{ghosh2023geneval} prompt dataset to evaluate complex text-to-image aligment ability.

\textbf{Verifier and Evaluation Metrics}: We used ImageReward \cite{xu2023imagereward} as the reward model, and assessed generation quality using PickScore~\cite{kirstain2023pick}, HPSv2~\cite{kirstain2023pick}, and the \textcolor{blue}{GenEva} metric~\cite{ghosh2023geneval}, Geneval is the most important and comprehensive benchmark on text-to-image aligment.

\textbf{Resampling Procedure}: For SMC-based methods, We adopted systematic resampling~\cite{chopin2020introduction} for $\mathcal{R}$, which reduces variance compared to multinomial resampling. The detailed difference is illustrated in Figure~\ref{fig:resampling_scheme} in Appendix~\ref{sec:appendix2}.\\

We conducted the following experiments to analyze the limitations of SMC in diffusion-based image generation and to evaluate the effectiveness of our proposed methods. Experiments involving SDXL were performed on an NVIDIA A100 GPU, while those for other models were carried out on an NVIDIA RTX 4090.

\textbf{Early-Stage Reward Correlation Analysis}: We analyze the correlation between intermediate and final reward rankings across different prompt categories in the GenEval dataset, highlighting the challenge of early-stage evaluation.
    
\textbf{Resampling Timing Analysis}: Using the DrawBench dataset, we evaluate the effect of resampling at different timesteps on final reward scores, revealing a stage-dependent impact of resampling on generation quality.
    
\textbf{Comprehensive Method Comparison}: We compare all baseline and proposed methods across multiple pre-trained models and evaluation metrics to assess their effectiveness and generalization.

\subsection{1: Analysis of Reward Reliability and Resampling Timing Across Diffusion Timesteps}
\label{sec:experiment1}

We analyzed the reliability of reward scores at different timesteps during the diffusion process using 100 randomly selected prompts from the DrawBench dataset. For each prompt, we sampled a batch of 16 noises and computed the Spearman rank correlation coefficient~\cite{dodge2008concise} between intermediate scores and the final score at timestep $t = 0$. The results, shown in Figure~\ref{fig:rank_correlation}, indicate that reward scores exhibit weak correlation with final image quality during the early stages. This contrasts with inverse tasks such as inpainting or super-resolution, which choose $\ell_2$-based metrics as the reward function.

We further investigate the effect of resampling timing in Figure~\ref{fig:influence}, where resampling is applied only once at a specified timestep. The results display a concave pattern: early-stage resampling has negligible or even negative impact on final quality, while resampling during the middle stages provides the greatest benefit, which is consistent with our discussion in Section~\ref{sec:general_search_framework}. After a certain transition point, the generation trajectory becomes semantically committed to a particular structure. Reward model can only guide the sampling process before this commitment.

\subsection{2: Performance Comparison with Different Methods}
\label{sec:experiment2}
We conducted experiments across various metrics, methods, and models. All methods were evaluated at denoising steps $\{10, 15, 20, 25, 29\}$, with the total number of noise function evaluations (NFE) fixed at 240, which means the particle count is $240 / 30 = 8$. our proposed \textbf{F-SMC} uses a dynamic particle schedule of $[12, 10, 6, 4, 4]$, which maintains the same total NFE.

For clarity, we denote methods using the \textit{Max} (as in FK-Steering~\cite{singhal2025generalframeworkinferencetimescaling}) with the suffix (M), and those using the \textit{Diff} with the suffix (D). The target reward temperature $\lambda$ is set to 10 for SD1.5, SD2.0, and SD2.1, and 6 for SDXL, due to its generally higher image quality, which necessitates a lower temperature to preserve diversity.

Results are summarized in Table~\ref{tab:inference_scaling}. Our proposed \textbf{F-SMC-A(D)} consistently outperforms all the methods and achieves the highest ImageReward (IR) and GenEval scores. We also compare the improvements of using the Funnel schedule in \textit{Max} potential or the adaptive temperature in \textit{Diff} potential individually in Table~\ref{tab:smc_comparison1} and Table~\ref{tab:smc_comparison2}. Interestingly, we observe that All the methods lead to a reduction in HPSv2 scores compared to the Best-of-N baseline. This highlights a key limitation of current reward models: strong alignment with one model does not necessarily imply better performance under others, pointing to the non-trivial generalization gap between reward objectives.

\begin{table}[t]
    \centering
    \begin{minipage}{0.48\textwidth}
        \centering
        \caption{Comparison of Using the F-SMC in \textit{Max} Potential}
        \label{tab:smc_comparison1}
        \begin{tabular}{l|l|c|c}
          \toprule
          \textbf{Method} & \textbf{Model} & \textbf{IR$\uparrow$} & \textbf{GenEval$\uparrow$} \\
          \midrule
          \multirow{4}{*}{SMC(M)}%
              & SD 1.5 & 1.000 & 55.52 \\
              & SD 2.0  & 1.197 & 61.12 \\
              & SD 2.1 & 1.240 & 60.04 \\
              & SD XL & 1.365 & 63.83 \\
          \midrule
          \multirow{4}{*}{F-SMC(M)}%
              & SD 1.5 & \textbf{1.016} & \textbf{55.70} \\
              & SD 2.0  & \textbf{1.189} & 60.22 \\
              & SD 2.1 & \textbf{1.270} & \textbf{60.94} \\
              & SD XL & \textbf{1.371} & \textbf{63.11} \\
          \bottomrule
        \end{tabular}
    \end{minipage}
    \hfill
    \begin{minipage}{0.48\textwidth}
        \centering
        \caption{Comparison of Using the SMC-A in \textit{Diff} Potential}
        \label{tab:smc_comparison2}
        \begin{tabular}{l|l|c|c}
          \toprule
          \textbf{Method} & \textbf{Model} & \textbf{IR$\uparrow$} & \textbf{GenEval$\uparrow$} \\
          \midrule
          \multirow{4}{*}{SMC(D)}%
              & SD 1.5 & 0.9403 & 54.97 \\
              & SD 2.0  & 1.158 & 56.24 \\
              & SD 2.1 & 1.222 & 60.94 \\
              & SD XL & 1.369 & 63.65 \\
          \midrule
          \multirow{4}{*}{SMC-A(D)}%
              & SD 1.5 & \textbf{1.007} & \textbf{56.78} \\
              & SD 2.0  & \textbf{1.249} & \textbf{63.29} \\
              & SD 2.1 & \textbf{1.288} & \textbf{61.66} \\
              & SD XL & \textbf{1.412} & \textbf{64.20} \\
          \bottomrule
        \end{tabular}
    \end{minipage}
\end{table}

\subsection{3: Performance with Scaling Inference Time}
\begin{figure}[t]
    \centering
    \includegraphics[width=1.0\linewidth]{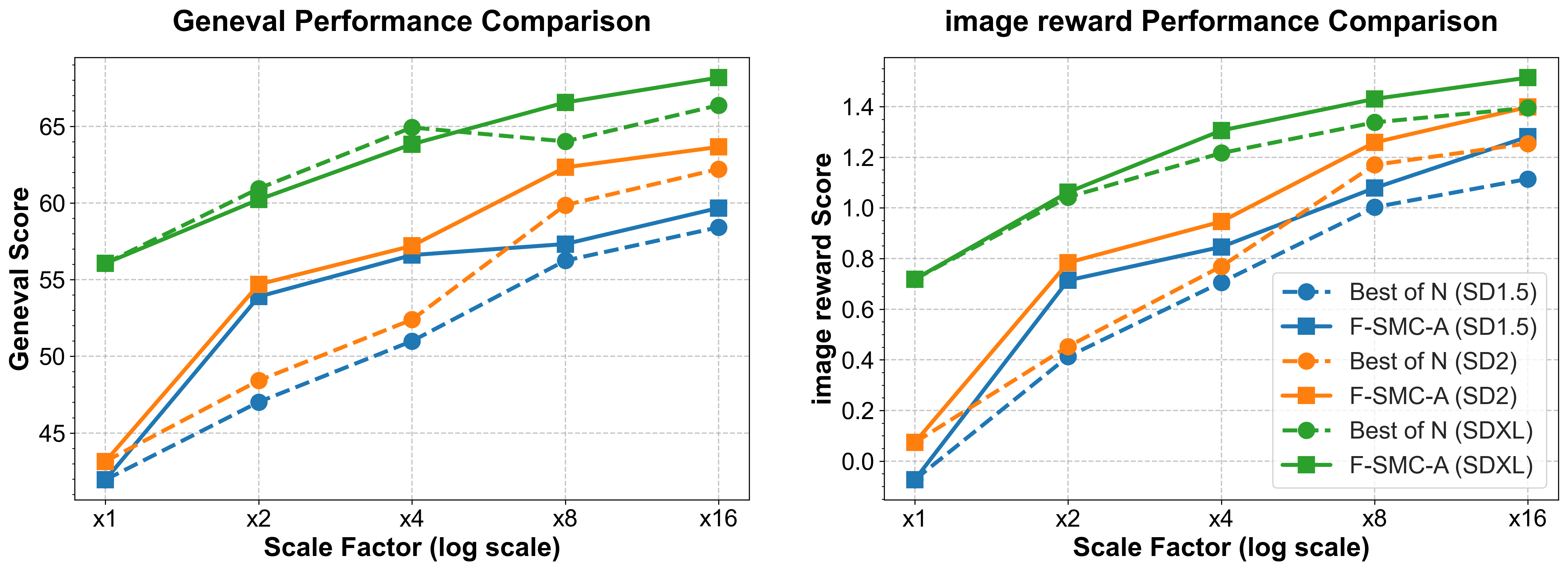}
    \caption{Effect of scaling on generation quality}
    \label{fig:scale}
\end{figure}

To evaluate how different methods benefit from increased computational resources, we analyzed their performance under varying inference-time budgets. The results are illustrated in Figure~\ref{fig:scale}, where the x-axis denotes the ratio of total Noise Function Evaluations (NFE) relative to a single generation pass. Inference-time scaling yields substantial improvements in text-to-image alignment across all methods. Notably, our proposed approach achieves comparable or superior performance to the Best-of-N baseline while requiring nearly half the NFE budget. This demonstrates the efficiency and scalability of F-SMC-A and its ability to make more effective use of limited sampling resources.

\section{Conclusion}
In this paper, we investigate inference-time scaling in DMs through the lens of search algorithms. We reveal two key challenges: the low reliability of reward estimators during early timesteps, and the progressive loss of malleability in the generation trajectory. To address these issues, we propose \textbf{F-SMC}, which allocates more particles to the early, more flexible stages of generation, and \textbf{adaptive temperature scheduling}, which compensates for early-stage reward inaccuracies. Both improvements preserve the theoretical convergence guarantees of standard SMC while delivering significantly better image-text alignment under a fixed computational budget. Looking forward, we aim to extend our methods to broader tasks such as aesthetic optimization and molecular generation. Additionally, we plan to improve reward models to better evaluate early-stage, ambiguous samples—further enhancing the robustness and effectiveness of inference-time search in DMs.

\bibliographystyle{plainnat}

\bibliography{references}

\begin{thebibliography}{59}
\providecommand{\natexlab}[1]{#1}
\providecommand{\url}[1]{\texttt{#1}}
\expandafter\ifx\csname urlstyle\endcsname\relax
  \providecommand{\doi}[1]{doi: #1}\else
  \providecommand{\doi}{doi: \begingroup \urlstyle{rm}\Url}\fi

\bibitem[BESKOS et~al.(2016)BESKOS, JASRA, KANTAS, and THIERY]{beskos2016convergence}
ALEXANDROS BESKOS, AJAY JASRA, NIKOLAS KANTAS, and ALEXANDRE THIERY.
\newblock On the convergence of adaptive sequential monte carlo methods.
\newblock \emph{The Annals of Applied Probability}, 26\penalty0 (2):\penalty0 1111--1146, 2016.

\bibitem[Chefer et~al.(2023)Chefer, Alaluf, Vinker, Wolf, and Cohen-Or]{chefer2023attend}
Hila Chefer, Yuval Alaluf, Yael Vinker, Lior Wolf, and Daniel Cohen-Or.
\newblock Attend-and-excite: Attention-based semantic guidance for text-to-image diffusion models.
\newblock \emph{ACM transactions on Graphics (TOG)}, 42\penalty0 (4):\penalty0 1--10, 2023.

\bibitem[Chen et~al.(2024{\natexlab{a}})Chen, Ge, Xie, Wu, Yao, Ren, Wang, Luo, Lu, and Li]{chen2024pixartsigmaweaktostrongtrainingdiffusion}
Junsong Chen, Chongjian Ge, Enze Xie, Yue Wu, Lewei Yao, Xiaozhe Ren, Zhongdao Wang, Ping Luo, Huchuan Lu, and Zhenguo Li.
\newblock Pixart-$\sigma$: Weak-to-strong training of diffusion transformer for 4k text-to-image generation.
\newblock In \emph{European Conference on Computer Vision}, pages 74--91, 2024{\natexlab{a}}.

\bibitem[Chen et~al.(2024{\natexlab{b}})Chen, Vaxman, Ben~Baruch, Asulin, Moreshet, Lien, Sra, and Sen]{chen2024tino}
Sherry~X Chen, Yaron Vaxman, Elad Ben~Baruch, David Asulin, Aviad Moreshet, Kuo-Chin Lien, Misha Sra, and Pradeep Sen.
\newblock Tino-edit: Timestep and noise optimization for robust diffusion-based image editing.
\newblock In \emph{Proceedings of the IEEE/CVF Conference on Computer Vision and Pattern Recognition}, pages 6337--6346, 2024{\natexlab{b}}.

\bibitem[Chen et~al.(2024{\natexlab{c}})Chen, Niu, Ma, Deng, Wang, Zhao, Yu, and Chen]{chen2024f5}
Yushen Chen, Zhikang Niu, Ziyang Ma, Keqi Deng, Chunhui Wang, Jian Zhao, Kai Yu, and Xie Chen.
\newblock F5-tts: A fairytaler that fakes fluent and faithful speech with flow matching.
\newblock \emph{arXiv:2410.06885}, 2024{\natexlab{c}}.

\bibitem[Chopin et~al.(2020)Chopin, Papaspiliopoulos, et~al.]{chopin2020introduction}
Nicolas Chopin, Omiros Papaspiliopoulos, et~al.
\newblock \emph{An introduction to sequential Monte Carlo}, volume~4.
\newblock Springer, 2020.

\bibitem[Chung et~al.(2023)Chung, Kim, Mccann, Klasky, and Ye]{chung2023diffusion}
Hyungjin Chung, Jeongsol Kim, Michael~Thompson Mccann, Marc~Louis Klasky, and Jong~Chul Ye.
\newblock Diffusion posterior sampling for general noisy inverse problems.
\newblock In \emph{The Eleventh International Conference on Learning Representations}, 2023.
\newblock URL \url{https://openreview.net/forum?id=OnD9zGAGT0k}.

\bibitem[Dieleman(2024)]{dieleman2024spectral}
Sander Dieleman.
\newblock Diffusion is spectral autoregression, 2024.
\newblock URL \url{https://sander.ai/2024/09/02/spectral-autoregression.html}.

\bibitem[Dodge(2008)]{dodge2008concise}
Yadolah Dodge.
\newblock \emph{The concise encyclopedia of statistics}.
\newblock Springer Science \& Business Media, 2008.

\bibitem[Dou and Song(2024)]{dou2024diffusion}
Zehao Dou and Yang Song.
\newblock Diffusion posterior sampling for linear inverse problem solving: A filtering perspective.
\newblock In \emph{The Twelfth International Conference on Learning Representations}, 2024.

\bibitem[Douc and Capp{\'e}(2005)]{douc2005comparison}
Randal Douc and Olivier Capp{\'e}.
\newblock Comparison of resampling schemes for particle filtering.
\newblock In \emph{ISPA 2005. Proceedings of the 4th International Symposium on Image and Signal Processing and Analysis, 2005.}, pages 64--69. Ieee, 2005.

\bibitem[Fan et~al.(2023)Fan, Watkins, Du, Liu, Ryu, Boutilier, Abbeel, Ghavamzadeh, Lee, and Lee]{fan2023dpok}
Ying Fan, Olivia Watkins, Yuqing Du, Hao Liu, Moonkyung Ryu, Craig Boutilier, Pieter Abbeel, Mohammad Ghavamzadeh, Kangwook Lee, and Kimin Lee.
\newblock Dpok: Reinforcement learning for fine-tuning text-to-image diffusion models.
\newblock \emph{Advances in Neural Information Processing Systems}, 36:\penalty0 79858--79885, 2023.

\bibitem[Gat et~al.(2024)Gat, Remez, Shaul, Kreuk, Chen, Synnaeve, Adi, and Lipman]{gat2024discrete}
Itai Gat, Tal Remez, Neta Shaul, Felix Kreuk, Ricky~TQ Chen, Gabriel Synnaeve, Yossi Adi, and Yaron Lipman.
\newblock Discrete flow matching.
\newblock \emph{Advances in Neural Information Processing Systems}, 37:\penalty0 133345--133385, 2024.

\bibitem[Ghosh et~al.(2023)Ghosh, Hajishirzi, and Schmidt]{ghosh2023geneval}
Dhruba Ghosh, Hannaneh Hajishirzi, and Ludwig Schmidt.
\newblock Geneval: An object-focused framework for evaluating text-to-image alignment.
\newblock \emph{Advances in Neural Information Processing Systems}, 36:\penalty0 52132--52152, 2023.

\bibitem[Guo et~al.(2024)Guo, Liu, Cui, Li, Yang, and Huang]{guo2024initno}
Xiefan Guo, Jinlin Liu, Miaomiao Cui, Jiankai Li, Hongyu Yang, and Di~Huang.
\newblock Initno: Boosting text-to-image diffusion models via initial noise optimization.
\newblock In \emph{Proceedings of the IEEE/CVF Conference on Computer Vision and Pattern Recognition}, pages 9380--9389, 2024.

\bibitem[Hao et~al.(2023)Hao, Gu, Ma, Hong, Wang, Wang, and Hu]{hao2023reasoning}
Shibo Hao, Yi~Gu, Haodi Ma, Joshua~Jiahua Hong, Zhen Wang, Daisy~Zhe Wang, and Zhiting Hu.
\newblock Reasoning with language model is planning with world model.
\newblock In \emph{The 2023 Conference on Empirical Methods in Natural Language Processing}, 2023.
\newblock URL \url{https://openreview.net/forum?id=VTWWvYtF1R}.

\bibitem[Ho et~al.(2020)Ho, Jain, and Abbeel]{ho2020denoisingdiffusionprobabilisticmodels}
Jonathan Ho, Ajay Jain, and Pieter Abbeel.
\newblock Denoising diffusion probabilistic models.
\newblock \emph{Advances in neural information processing systems}, 33:\penalty0 6840--6851, 2020.

\bibitem[Ho et~al.(2022)Ho, Salimans, Gritsenko, Chan, Norouzi, and Fleet]{ho2022video}
Jonathan Ho, Tim Salimans, Alexey Gritsenko, William Chan, Mohammad Norouzi, and David~J Fleet.
\newblock Video diffusion models.
\newblock \emph{Advances in Neural Information Processing Systems}, 35:\penalty0 8633--8646, 2022.

\bibitem[Hodaň et~al.(2019)Hodaň, Vineet, Gal, Shalev, Hanzelka, Connell, Urbina, Sinha, and Guenter]{8803821}
Tomáš Hodaň, Vibhav Vineet, Ran Gal, Emanuel Shalev, Jon Hanzelka, Treb Connell, Pedro Urbina, Sudipta~N. Sinha, and Brian Guenter.
\newblock Photorealistic image synthesis for object instance detection.
\newblock In \emph{2019 IEEE International Conference on Image Processing (ICIP)}, pages 66--70, 2019.
\newblock \doi{10.1109/ICIP.2019.8803821}.

\bibitem[Jeong et~al.(2021)Jeong, Kim, Cheon, Choi, and Kim]{jeong2021diff}
Myeonghun Jeong, Hyeongju Kim, Sung~Jun Cheon, Byoung~Jin Choi, and Nam~Soo Kim.
\newblock Diff-tts: A denoising diffusion model for text-to-speech.
\newblock In \emph{22nd Annual Conference of the International Speech Communication Association, INTERSPEECH 2021}, pages 3566--3570. International Speech Communication Association, 2021.

\bibitem[Kadkhodaie and Simoncelli(2021)]{kadkhodaie2021stochastic}
Zahra Kadkhodaie and Eero Simoncelli.
\newblock Stochastic solutions for linear inverse problems using the prior implicit in a denoiser.
\newblock \emph{Advances in Neural Information Processing Systems}, 34:\penalty0 13242--13254, 2021.

\bibitem[Kaplan et~al.(2020)Kaplan, McCandlish, Henighan, Brown, Chess, Child, Gray, Radford, Wu, and Amodei]{kaplan2020scalinglawsneurallanguage}
Jared Kaplan, Sam McCandlish, Tom Henighan, Tom~B. Brown, Benjamin Chess, Rewon Child, Scott Gray, Alec Radford, Jeffrey Wu, and Dario Amodei.
\newblock Scaling laws for neural language models, 2020.
\newblock URL \url{https://arxiv.org/abs/2001.08361}.

\bibitem[Kim et~al.(2025{\natexlab{a}})Kim, Yoon, Hwang, and Sung]{kim2025inferencetimescalingflowmodels}
Jaihoon Kim, Taehoon Yoon, Jisung Hwang, and Minhyuk Sung.
\newblock Inference-time scaling for flow models via stochastic generation and rollover budget forcing, 2025{\natexlab{a}}.
\newblock URL \url{https://arxiv.org/abs/2503.19385}.

\bibitem[Kim et~al.(2025{\natexlab{b}})Kim, Park, Chung, and Ye]{kim2025regularization}
Jeongsol Kim, Geon~Yeong Park, Hyungjin Chung, and Jong~Chul Ye.
\newblock Regularization by texts for latent diffusion inverse solvers.
\newblock In \emph{The Thirteenth International Conference on Learning Representations}, 2025{\natexlab{b}}.
\newblock URL \url{https://openreview.net/forum?id=TtUh0TOlGX}.

\bibitem[Kirstain et~al.(2023)Kirstain, Polyak, Singer, Matiana, Penna, and Levy]{kirstain2023pick}
Yuval Kirstain, Adam Polyak, Uriel Singer, Shahbuland Matiana, Joe Penna, and Omer Levy.
\newblock Pick-a-pic: An open dataset of user preferences for text-to-image generation.
\newblock \emph{Advances in Neural Information Processing Systems}, 36:\penalty0 36652--36663, 2023.

\bibitem[Li and Chen(2024)]{li2024criticalwindows}
Marvin Li and Sitan Chen.
\newblock Critical windows: non-asymptotic theory for feature emergence in diffusion models.
\newblock In \emph{Proceedings of the 41st International Conference on Machine Learning}, ICML'24. JMLR.org, 2024.

\bibitem[Li et~al.(2024)Li, Zhao, Wang, Scalia, Eraslan, Nair, Biancalani, Ji, Regev, Levine, and Uehara]{li2024derivativefreeguidancecontinuousdiscrete}
Xiner Li, Yulai Zhao, Chenyu Wang, Gabriele Scalia, Gokcen Eraslan, Surag Nair, Tommaso Biancalani, Shuiwang Ji, Aviv Regev, Sergey Levine, and Masatoshi Uehara.
\newblock Derivative-free guidance in continuous and discrete diffusion models with soft value-based decoding, 2024.
\newblock URL \url{https://arxiv.org/abs/2408.08252}.

\bibitem[LiChen et~al.(2024)LiChen, Shao, Qi, Xiong, Xie, et~al.]{bai2024zigzagdiffusionsamplingdiffusion}
Bai LiChen, Shitong Shao, Zipeng Qi, Haoyi Xiong, Zeke Xie, et~al.
\newblock Zigzag diffusion sampling: Diffusion models can self-improve via self-reflection.
\newblock In \emph{The Thirteenth International Conference on Learning Representations}, 2024.

\bibitem[Liu et~al.(2025)Liu, Liu, Liang, Li, Liu, Wang, Wan, Zhang, and Ouyang]{liu2025flowgrpotrainingflowmatching}
Jie Liu, Gongye Liu, Jiajun Liang, Yangguang Li, Jiaheng Liu, Xintao Wang, Pengfei Wan, Di~Zhang, and Wanli Ouyang.
\newblock Flow-grpo: Training flow matching models via online rl, 2025.
\newblock URL \url{https://arxiv.org/abs/2505.05470}.

\bibitem[Loula et~al.(2025)Loula, LeBrun, Du, Lipkin, Pasti, Grand, Liu, Emara, Freedman, Eisner, Cotterell, Mansinghka, Lew, Vieira, and O'Donnell]{loula2025syntactic}
Jo{\~a}o Loula, Benjamin LeBrun, Li~Du, Ben Lipkin, Clemente Pasti, Gabriel Grand, Tianyu Liu, Yahya Emara, Marjorie Freedman, Jason Eisner, Ryan Cotterell, Vikash Mansinghka, Alexander~K. Lew, Tim Vieira, and Timothy~J. O'Donnell.
\newblock Syntactic and semantic control of large language models via sequential monte carlo.
\newblock In \emph{The Thirteenth International Conference on Learning Representations}, 2025.
\newblock URL \url{https://openreview.net/forum?id=xoXn62FzD0}.

\bibitem[Lu et~al.(2024)Lu, Xu, Zhang, Wang, and Tao]{lu2024handrefiner}
Wenquan Lu, Yufei Xu, Jing Zhang, Chaoyue Wang, and Dacheng Tao.
\newblock Handrefiner: Refining malformed hands in generated images by diffusion-based conditional inpainting.
\newblock In \emph{Proceedings of the 32nd ACM International Conference on Multimedia}, MM '24, page 7085–7093, New York, NY, USA, 2024. Association for Computing Machinery.
\newblock ISBN 9798400706868.
\newblock \doi{10.1145/3664647.3680693}.
\newblock URL \url{https://doi.org/10.1145/3664647.3680693}.

\bibitem[Ma et~al.(2025)Ma, Tong, Jia, Hu, Su, Zhang, Yang, Li, Jaakkola, Jia, and Xie]{ma2025inference}
Nanye Ma, Shangyuan Tong, Haolin Jia, Hexiang Hu, Yu-Chuan Su, Mingda Zhang, Xuan Yang, Yandong Li, Tommi Jaakkola, Xuhui Jia, and Saining Xie.
\newblock Scaling inference time compute for diffusion models.
\newblock In \emph{Proceedings of the IEEE/CVF Conference on Computer Vision and Pattern Recognition (CVPR)}, pages 2523--2534, June 2025.

\bibitem[Mardani et~al.(2023)Mardani, Song, Kautz, and Vahdat]{mardani2023variational}
Morteza Mardani, Jiaming Song, Jan Kautz, and Arash Vahdat.
\newblock A variational perspective on solving inverse problems with diffusion models.
\newblock In \emph{The Twelfth International Conference on Learning Representations}, 2023.

\bibitem[Mardani et~al.(2024)Mardani, Song, Kautz, and Vahdat]{mardani2024a}
Morteza Mardani, Jiaming Song, Jan Kautz, and Arash Vahdat.
\newblock A variational perspective on solving inverse problems with diffusion models.
\newblock In \emph{The Twelfth International Conference on Learning Representations}, 2024.
\newblock URL \url{https://openreview.net/forum?id=1YO4EE3SPB}.

\bibitem[Moore(1959)]{moore1959shortest}
Edward~F. Moore.
\newblock The shortest path through a maze.
\newblock In \emph{Proceedings of an International Symposium on the Theory of Switching, Part II}, pages 285--292, Cambridge, MA, 1959. Harvard University Press.

\bibitem[Nie et~al.(2025)Nie, Zhu, You, Zhang, Ou, Hu, ZHOU, Lin, Wen, and Li]{nie2025largelanguagediffusionmodels}
Shen Nie, Fengqi Zhu, Zebin You, Xiaolu Zhang, Jingyang Ou, Jun Hu, JUN ZHOU, Yankai Lin, Ji-Rong Wen, and Chongxuan Li.
\newblock Large language diffusion models.
\newblock In \emph{ICLR Workshop on Deep Generative Model in Machine Learning: Theory, Principle and Efficacy}, 2025.

\bibitem[Pavasovic et~al.(2025)Pavasovic, Verbeek, Biroli, and Mezard]{pavasovic2025understandingclassifierfreeguidancehighdimensional}
Krunoslav~Lehman Pavasovic, Jakob Verbeek, Giulio Biroli, and Marc Mezard.
\newblock Understanding classifier-free guidance: High-dimensional theory and non-linear generalizations, 2025.
\newblock URL \url{https://arxiv.org/abs/2502.07849}.

\bibitem[Podell et~al.(2024)Podell, English, Lacey, Blattmann, Dockhorn, M{\"u}ller, Penna, and Rombach]{podell2024sdxl}
Dustin Podell, Zion English, Kyle Lacey, Andreas Blattmann, Tim Dockhorn, Jonas M{\"u}ller, Joe Penna, and Robin Rombach.
\newblock {SDXL}: Improving latent diffusion models for high-resolution image synthesis.
\newblock In \emph{The Twelfth International Conference on Learning Representations}, 2024.
\newblock URL \url{https://openreview.net/forum?id=di52zR8xgf}.

\bibitem[Poole et~al.(2023)Poole, Jain, Barron, and Mildenhall]{poole2022dreamfusion}
Ben Poole, Ajay Jain, Jonathan~T Barron, and Ben Mildenhall.
\newblock Dreamfusion: Text-to-3d using 2d diffusion.
\newblock In \emph{The Eleventh International Conference on Learning Representations}, 2023.

\bibitem[Qi et~al.(2024)Qi, Bai, Xiong, and Xie]{qi2024noisescreatedequallydiffusionnoise}
Zipeng Qi, Lichen Bai, Haoyi Xiong, and Zeke Xie.
\newblock Not all noises are created equally:diffusion noise selection and optimization, 2024.
\newblock URL \url{https://arxiv.org/abs/2407.14041}.

\bibitem[Qiu et~al.(2024)Qiu, Lu, Zeng, Guo, Geng, Wang, Huang, Wu, and Wang]{qiu2024treebon}
Jiahao Qiu, Yifu Lu, Yifan Zeng, Jiacheng Guo, Jiayi Geng, Huazheng Wang, Kaixuan Huang, Yue Wu, and Mengdi Wang.
\newblock Treebon: Enhancing inference-time alignment with speculative tree-search and best-of-n sampling.
\newblock \emph{arXiv:2410.16033}, 2024.

\bibitem[Rombach et~al.(2022)Rombach, Blattmann, Lorenz, Esser, and Ommer]{rombach2022highresolutionimagesynthesislatent}
Robin Rombach, Andreas Blattmann, Dominik Lorenz, Patrick Esser, and Bjorn Ommer.
\newblock High-resolution image synthesis with latent diffusion models.
\newblock In \emph{2022 IEEE/CVF Conference on Computer Vision and Pattern Recognition (CVPR)}, pages 10674--10685. IEEE Computer Society, 2022.

\bibitem[Singhal et~al.(2025)Singhal, Horvitz, Teehan, Ren, Yu, McKeown, and Ranganath]{singhal2025generalframeworkinferencetimescaling}
Raghav Singhal, Zachary Horvitz, Ryan Teehan, Mengye Ren, Zhou Yu, Kathleen McKeown, and Rajesh Ranganath.
\newblock A general framework for inference-time scaling and steering of diffusion models.
\newblock In \emph{Forty-second International Conference on Machine Learning}, 2025.

\bibitem[Sohl-Dickstein et~al.(2015)Sohl-Dickstein, Weiss, Maheswaranathan, and Ganguli]{pmlr-v37-sohl-dickstein15}
Jascha Sohl-Dickstein, Eric Weiss, Niru Maheswaranathan, and Surya Ganguli.
\newblock Deep unsupervised learning using nonequilibrium thermodynamics.
\newblock In Francis Bach and David Blei, editors, \emph{Proceedings of the 32nd International Conference on Machine Learning}, volume~37 of \emph{Proceedings of Machine Learning Research}, pages 2256--2265, Lille, France, 07--09 Jul 2015. PMLR.
\newblock URL \url{https://proceedings.mlr.press/v37/sohl-dickstein15.html}.

\bibitem[Song et~al.(2023)Song, Vahdat, Mardani, and Kautz]{song2023pseudoinverseguided}
Jiaming Song, Arash Vahdat, Morteza Mardani, and Jan Kautz.
\newblock Pseudoinverse-guided diffusion models for inverse problems.
\newblock In \emph{International Conference on Learning Representations}, 2023.
\newblock URL \url{https://openreview.net/forum?id=9_gsMA8MRKQ}.

\bibitem[Song et~al.(2021)Song, Sohl-Dickstein, Kingma, Kumar, Ermon, and Poole]{song2021scorebasedgenerativemodelingstochastic}
Yang Song, Jascha Sohl-Dickstein, Diederik~P Kingma, Abhishek Kumar, Stefano Ermon, and Ben Poole.
\newblock Score-based generative modeling through stochastic differential equations.
\newblock In \emph{International Conference on Learning Representations}, 2021.

\bibitem[Tarjan(1972)]{tarjan1972dfs}
Robert~E. Tarjan.
\newblock Depth‑first search and linear graph algorithms.
\newblock \emph{SIAM Journal on Computing}, 1\penalty0 (2):\penalty0 146--160, 1972.

\bibitem[Wallace et~al.(2024)Wallace, Dang, Rafailov, Zhou, Lou, Purushwalkam, Ermon, Xiong, Joty, and Naik]{wallace2024diffusion}
Bram Wallace, Meihua Dang, Rafael Rafailov, Linqi Zhou, Aaron Lou, Senthil Purushwalkam, Stefano Ermon, Caiming Xiong, Shafiq Joty, and Nikhil Naik.
\newblock Diffusion model alignment using direct preference optimization.
\newblock In \emph{Proceedings of the IEEE/CVF Conference on Computer Vision and Pattern Recognition}, pages 8228--8238, 2024.

\bibitem[Wang et~al.(2023{\natexlab{a}})Wang, Wei, Schuurmans, Le, Chi, Narang, Chowdhery, and Zhou]{wangself}
Xuezhi Wang, Jason Wei, Dale Schuurmans, Quoc~V Le, Ed~H Chi, Sharan Narang, Aakanksha Chowdhery, and Denny Zhou.
\newblock Self-consistency improves chain of thought reasoning in language models.
\newblock In \emph{The Eleventh International Conference on Learning Representations}, 2023{\natexlab{a}}.

\bibitem[Wang et~al.(2022)Wang, Yu, and Zhang]{wang2022zero}
Yinhuai Wang, Jiwen Yu, and Jian Zhang.
\newblock Zero-shot image restoration using denoising diffusion null-space model.
\newblock \emph{arXiv preprint arXiv:2212.00490}, 2022.

\bibitem[Wang et~al.(2023{\natexlab{b}})Wang, Yu, and Zhang]{wang2023zeroshot}
Yinhuai Wang, Jiwen Yu, and Jian Zhang.
\newblock Zero-shot image restoration using denoising diffusion null-space model.
\newblock In \emph{The Eleventh International Conference on Learning Representations}, 2023{\natexlab{b}}.
\newblock URL \url{https://openreview.net/forum?id=mRieQgMtNTQ}.

\bibitem[Wu et~al.(2023)Wu, Trippe, Naesseth, Blei, and Cunningham]{wu2023practical}
Luhuan Wu, Brian Trippe, Christian Naesseth, David Blei, and John~P Cunningham.
\newblock Practical and asymptotically exact conditional sampling in diffusion models.
\newblock \emph{Advances in Neural Information Processing Systems}, 36:\penalty0 31372--31403, 2023.

\bibitem[Wu et~al.(2025)Wu, Sun, Li, Welleck, and Yang]{wu2025inference}
Yangzhen Wu, Zhiqing Sun, Shanda Li, Sean Welleck, and Yiming Yang.
\newblock Inference scaling laws: An empirical analysis of compute-optimal inference for llm problem-solving.
\newblock In \emph{The Thirteenth International Conference on Learning Representations}, 2025.

\bibitem[Xie et~al.(2025)Xie, Chen, Zhao, YU, Zhu, Lin, Zhang, Li, Chen, Cai, et~al.]{xie2025sana15efficientscaling}
Enze Xie, Junsong Chen, Yuyang Zhao, Jincheng YU, Ligeng Zhu, Yujun Lin, Zhekai Zhang, Muyang Li, Junyu Chen, Han Cai, et~al.
\newblock Sana 1.5: Efficient scaling of training-time and inference-time compute in linear diffusion transformer.
\newblock In \emph{Forty-second International Conference on Machine Learning}, 2025.

\bibitem[Xu et~al.(2023)Xu, Liu, Wu, Tong, Li, Ding, Tang, and Dong]{xu2023imagereward}
Jiazheng Xu, Xiao Liu, Yuchen Wu, Yuxuan Tong, Qinkai Li, Ming Ding, Jie Tang, and Yuxiao Dong.
\newblock Imagereward: Learning and evaluating human preferences for text-to-image generation.
\newblock \emph{Advances in Neural Information Processing Systems}, 36:\penalty0 15903--15935, 2023.

\bibitem[Yao et~al.(2023)Yao, Yu, Zhao, Shafran, Griffiths, Cao, and Narasimhan]{NEURIPS2023_271db992}
Shunyu Yao, Dian Yu, Jeffrey Zhao, Izhak Shafran, Tom Griffiths, Yuan Cao, and Karthik Narasimhan.
\newblock Tree of thoughts: Deliberate problem solving with large language models.
\newblock In A.~Oh, T.~Naumann, A.~Globerson, K.~Saenko, M.~Hardt, and S.~Levine, editors, \emph{Advances in Neural Information Processing Systems}, volume~36, pages 11809--11822. Curran Associates, Inc., 2023.
\newblock URL \url{https://proceedings.neurips.cc/paper_files/paper/2023/file/271db9922b8d1f4dd7aaef84ed5ac703-Paper-Conference.pdf}.

\bibitem[Yu et~al.(2023)Yu, Wang, Zhao, Ghanem, and Zhang]{yu2023freedom}
Jiwen Yu, Yinhuai Wang, Chen Zhao, Bernard Ghanem, and Jian Zhang.
\newblock Freedom: Training-free energy-guided conditional diffusion model.
\newblock In \emph{Proceedings of the IEEE/CVF International Conference on Computer Vision}, pages 23174--23184, 2023.

\bibitem[Zhang et~al.(2025)Zhang, Lin, Ye, Zou, Ma, Liang, and Du]{zhang2025inferencetimescalingdiffusionmodels}
Xiangcheng Zhang, Haowei Lin, Haotian Ye, James Zou, Jianzhu Ma, Yitao Liang, and Yilun Du.
\newblock Inference-time scaling of diffusion models through classical search, 2025.
\newblock URL \url{https://arxiv.org/abs/2505.23614}.

\bibitem[Zhao et~al.(2023)Zhao, Lee, and Hsu]{zhao2023large}
Zirui Zhao, Wee~Sun Lee, and David Hsu.
\newblock Large language models as commonsense knowledge for large-scale task planning.
\newblock In \emph{RSS 2023 Workshop on Learning for Task and Motion Planning}, 2023.
\newblock URL \url{https://openreview.net/forum?id=tED747HURfX}.

\end{thebibliography}


\clearpage
\appendix
\section{Appendix}

\subsection{1: Illustration of Search Algorithms}

We consider a DM as a generative prior for producing blue squares on a checkerboard. As shown in Figure \ref{fig:method}, a Gaussian mixture distribution is used as the reward distribution, assigning different probabilities to each blue square on the board. This setup represents a multimodal search problem. In the early stages, prior to the phase transition, particles remain uncertain about which square they will ultimately occupy. Once a square is selected, it becomes difficult to compel particles to transition to another. An effective search method for DMs must therefore support both global and local search capabilities.

\subsection{2: Resampling Method}
\label{sec:appendix2}

There are two common resampling schemes in SMC methods: multinomial resampling and systematic resampling \cite{douc2005comparison}, as illustrated in Figure~\ref{fig:resampling_scheme}. Multinomial resampling randomly selects particles from the current population with replacement, whereas systematic resampling selects a random starting point in the interval $[0, 1/N)$ and proceeds with fixed step sizes of $1/N$. Systematic resampling offers better sample diversity and lower variance than multinomial resampling, particularly when the number of particles is small. Owing to these advantages, systematic resampling has become the de facto standard in most modern SMC implementations.

\begin{figure}[t]
\centering
\includegraphics[width=0.5\linewidth]{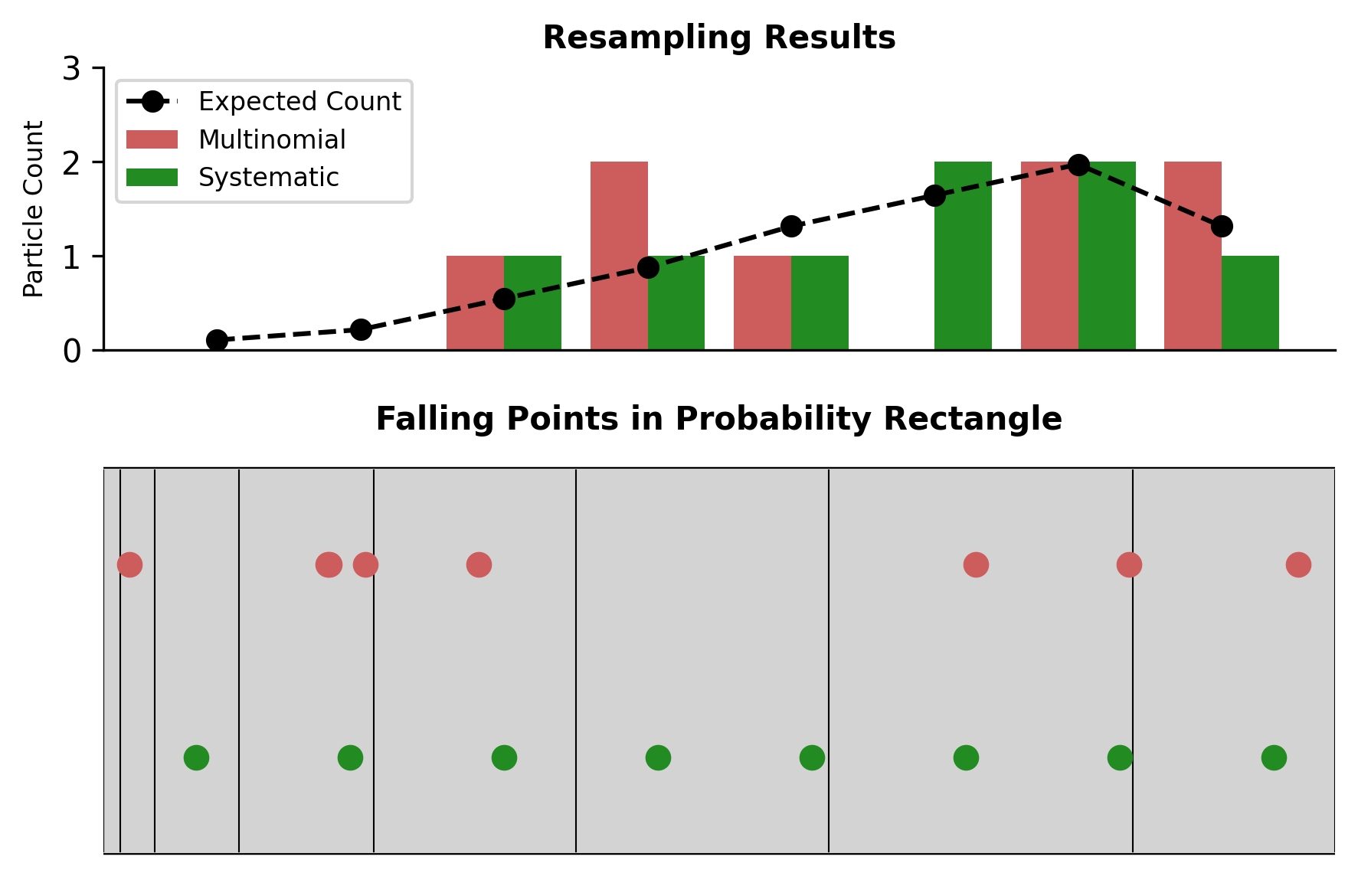}
\caption{Comparison of multinomial resampling and systematic resampling}
\label{fig:resampling_scheme}
\end{figure}

\subsection{3: Early-stage Reward Correlation Analysis}
\label{sec:appendix3}

\begin{figure}[t]
    \centering
    \begin{minipage}{0.48\textwidth}
        \centering
        \includegraphics[width=\textwidth]{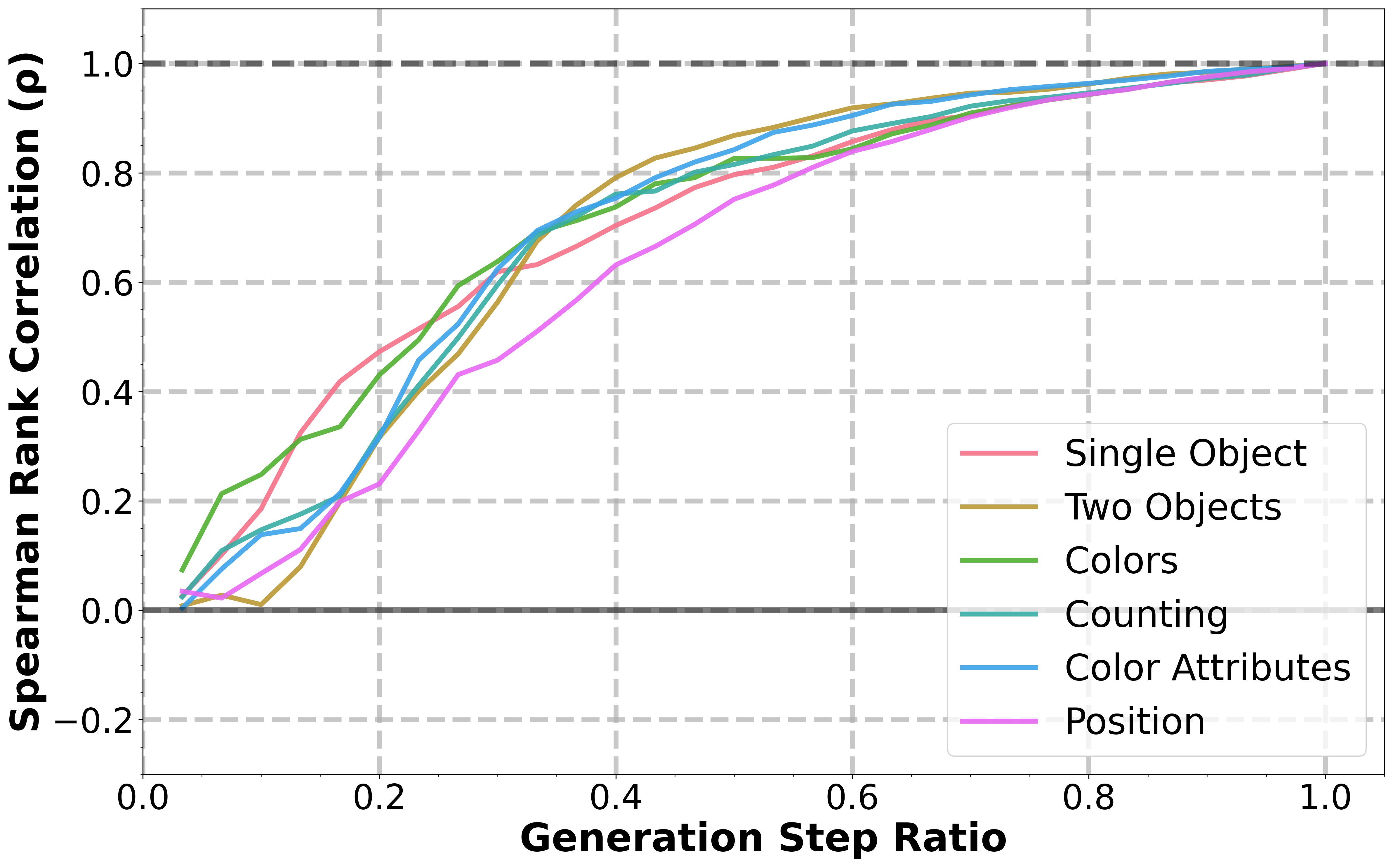}
        \caption{Correlation analysis for SD 2.0}
        \label{fig:tag_correlations_sd2}
    \end{minipage}
    \hfill
    \begin{minipage}{0.48\textwidth}
        \centering
        \includegraphics[width=\textwidth]{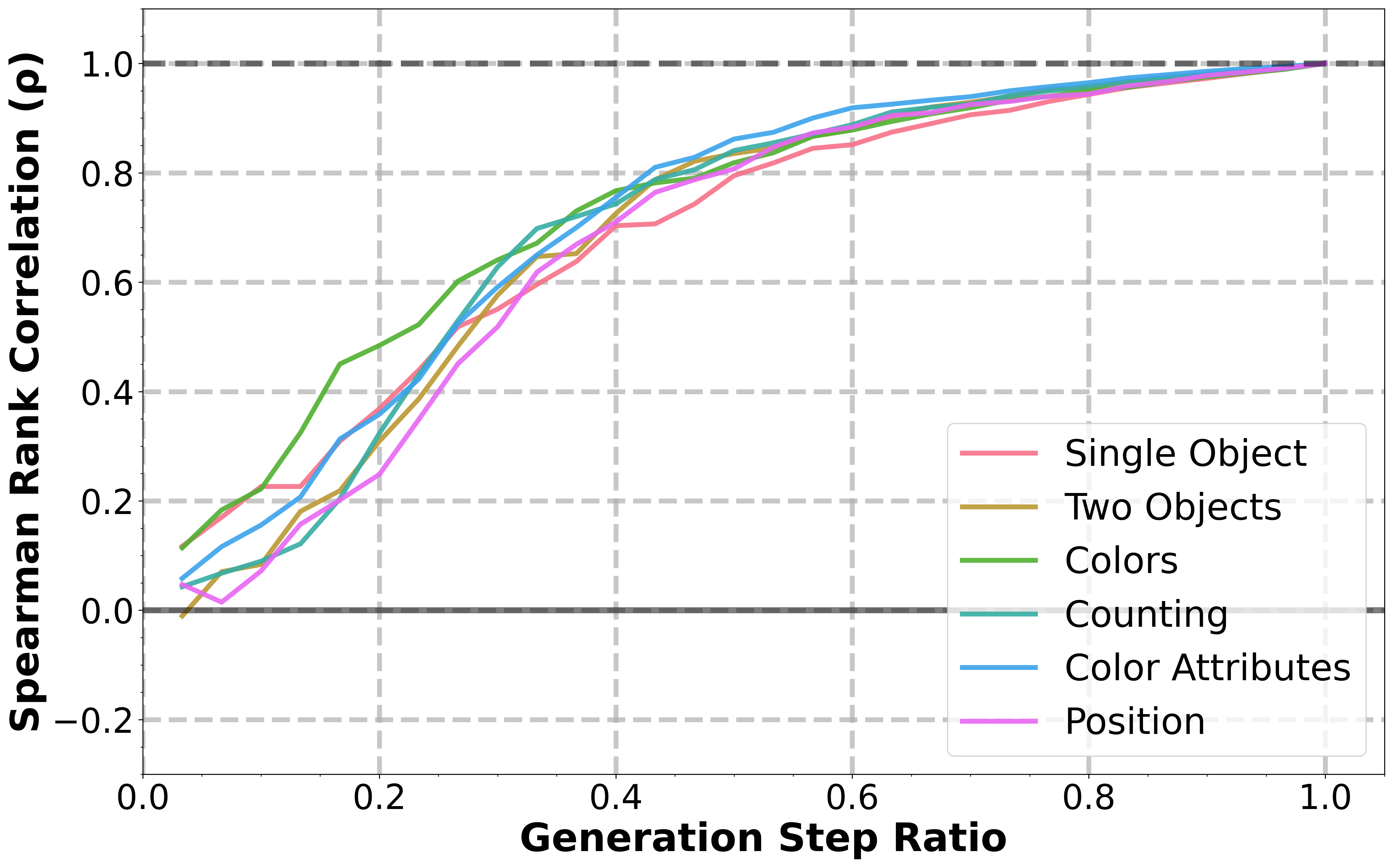}
        \caption{Correlation analysis for SD 1.5}
        \label{fig:tag_correlations_sd15}
    \end{minipage}
\end{figure}

The accuracy of early-stage evaluations by the reward model is critical for prompt quality. We assess the model's accuracy in early evaluations across different prompt types in the GenEval dataset. Using the same method described in Section~\ref{sec:experiment1}, we plot the progression of Spearman's rank correlation coefficient over the generation steps in Figure~\ref{fig:tag_correlations_sd2} and Figure~\ref{fig:tag_correlations_sd15}. It is evident that the correlation for the \textit{position} category remains consistently low, which leads to the poor performance of image-reward-based models on this category in the GenEval test (averaging below 20\%).

\subsection{4: Influence of Funnel Schedule}
\label{sec:appendix4}

\begin{figure}[t]
    \centering
    \begin{minipage}{0.48\textwidth}
        \centering
        \includegraphics[width=\textwidth]{figures/resampling_scheme.png}
        \caption{Comparison of multinomial resampling and systematic resampling}
        \label{fig:resampling_scheme}
    \end{minipage}
    \hfill
    \begin{minipage}{0.48\textwidth}
        \centering
        \includegraphics[width=\textwidth]{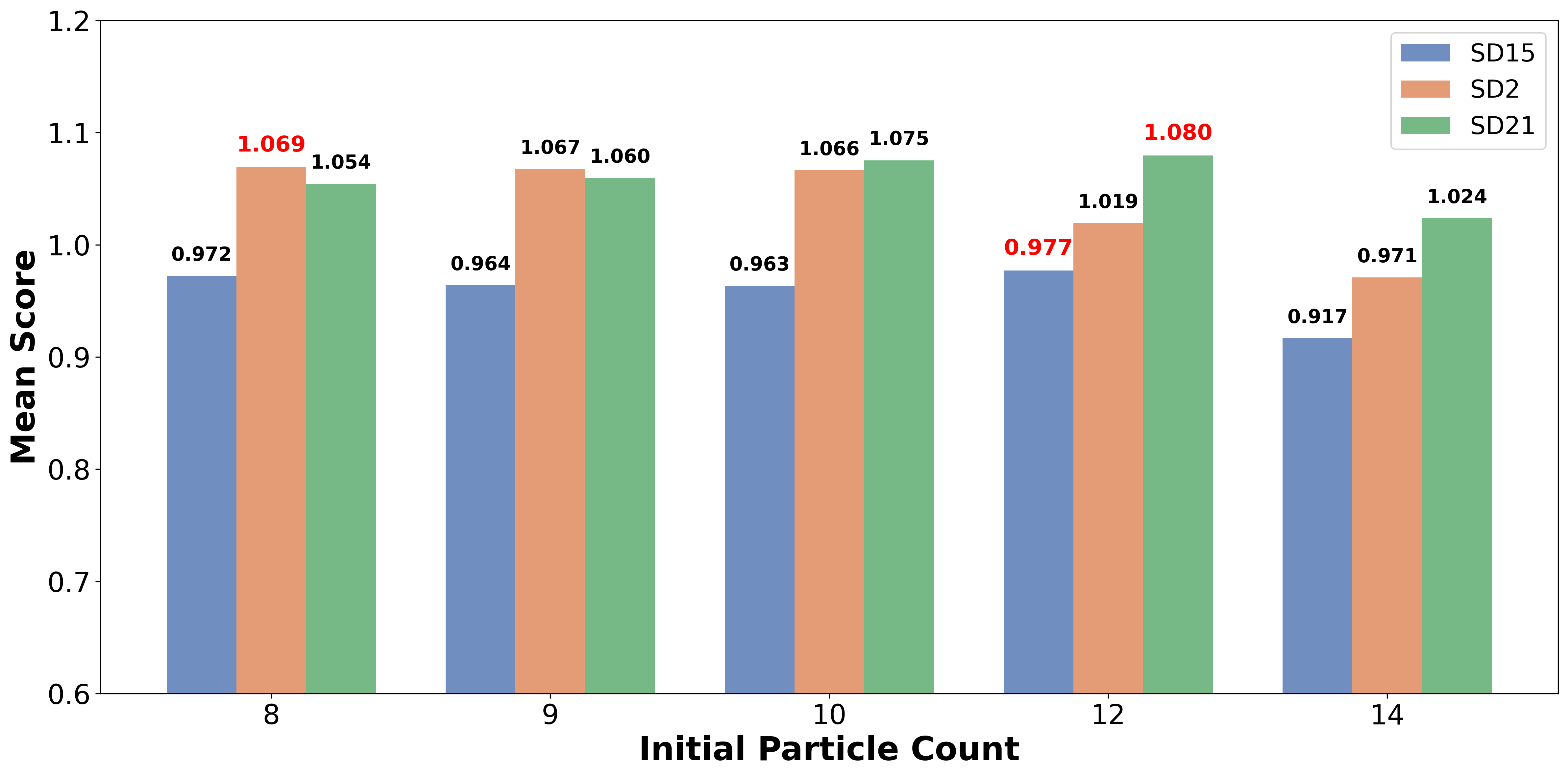}
        \caption{Comparison of ImageReward scores of different funnel schedules}
        \label{fig:aggressive_comparison}
    \end{minipage}
\end{figure}

To investigate the effect of different funnel schedules, we tested several models on DrawBench while keeping the total number of function evaluations (NFE) fixed at 240, and resampling at the same steps as in Section~\ref{sec:experiment2}, specifically at steps [10, 15, 20, 25, 29]. The following series of funnel schedules were evaluated:
\begin{itemize}
    \item Schedule 1: [8, 8, 8, 8, 8]
    \item Schedule 2: [9, 8, 8, 7, 7]
    \item Schedule 3: [10, 8, 8, 6, 6]
    \item Schedule 4: [12, 10, 6, 4, 4]
    \item Schedule 5: [14, 10, 6, 2, 2]
\end{itemize}
The results for models SD1.5, SD2.0, and SD2.1 are presented in Figure~\ref{fig:aggressive_comparison}. We find that selecting 12 particles,corresponding to Schedule 4, which we currently use, yields the best performance. Accordingly, this schedule will be used for subsequent experiments on the GenEval dataset.

\subsection{5: Convergence of F-SMC-A}
\label{sec:appendix3}

We have shown that, under the resampling schedule of the F-SMC-A algorithm, choosing either the \textit{Max} or \textit{Diff} potential still ensures that the particle weights at time step $t=0$ are proportional to those of the target distribution (see Proof~\ref{proof: potential}). What remains is to establish the convergence of the F-SMC-A algorithm under dynamic conditions, where both the number of particles $N_t$ and the adaptive temperature $\lambda_t$ vary over time. Specifically, we aim to show that the F-SMC-A algorithm converges to the target distribution in this time-varying setting.

This convergence problem has been previously studied in the SMC literature~\cite{beskos2016convergence}. The convergence of such adaptive SMC methods is guaranteed, provided that certain additional conditions are satisfied beyond those required by standard SMC algorithms:

\begin{itemize}
    \item \textbf{Bounded potentials.}\;
          \(G_{t}=\exp[\lambda_t r_t]\)
          is strictly positive and bounded, uniformly in \(t\).
    \item \textbf{Smooth dependence on \(\lambda_t\).}\;
          For each \(t\) the mapping
          \(\lambda\mapsto G_{t}(\vec{x}_{T:t}, \vec{c}, \lambda_t)\) is continuous with a
          bounded first derivative, uniformly in \(\vec{x}_t\).
    \item \textbf{Diverging particles.}\;
          For every fixed time index \(t\) we have \(N_t\to\infty\).
\end{itemize}

The former two conditions are satisfied by the proposed \textit{Max} potential and \textit{Diff} potential. The last condition is satisfied by the dynamic particle count $N_t$ because we use a non-increasing sequence
\(N_0\leq N_1\leq\cdots\leq N_T\). And we make sure that the particle ratio \(N_0 / N_t > 0\) for all \(t\).

\subsection{6: Comparative Illustration of Proposed and Existing Methods}

This section presents a comparative analysis between our proposed method and existing approaches. We highlight two representative scenarios where our method demonstrates superior performance.

\begin{figure*}[h]
    \centering
    \caption{Global Improvement: Comparison of Proposed Method with Baseline Methods}
    \label{fig:method_comparison_global}
    Best-of-N \hspace{2.5cm}  SMC(M) \hspace{2.5cm} F-SMC(M) \hspace{2.5cm} F-SMC-A(D)\\
    \includegraphics[width=\textwidth]{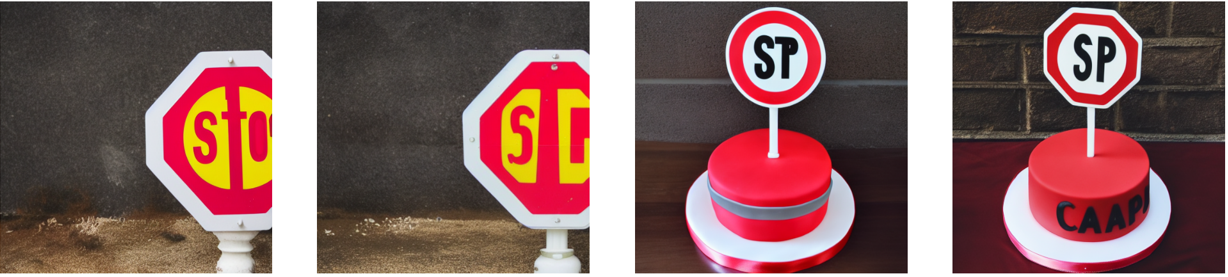}\\
    Prompt: a photo of a cake and a stop sign. Model: SD 1.5
    \includegraphics[width=\textwidth]{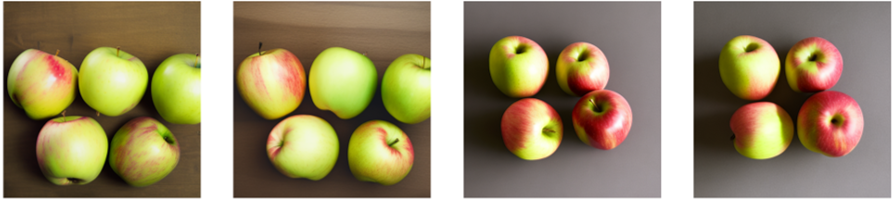}\\
    Prompt: a photo of four apples. Model: SD 2.0
    \includegraphics[width=\textwidth]{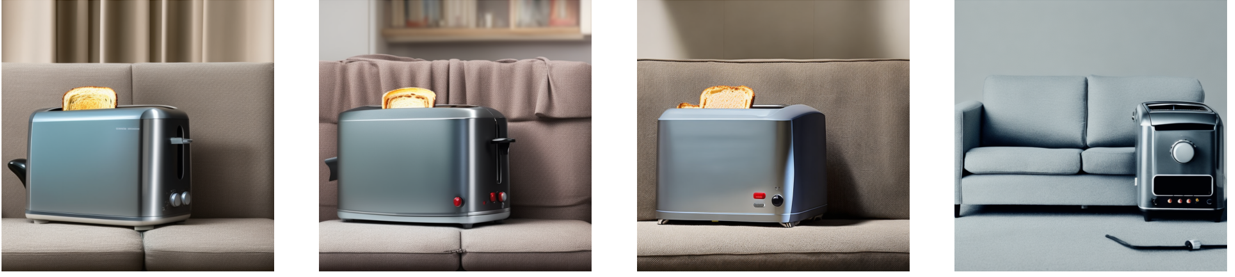}\\
    Prompt: a photo of a couch left of a toaster. Model: SD 2.1
    \includegraphics[width=\textwidth]{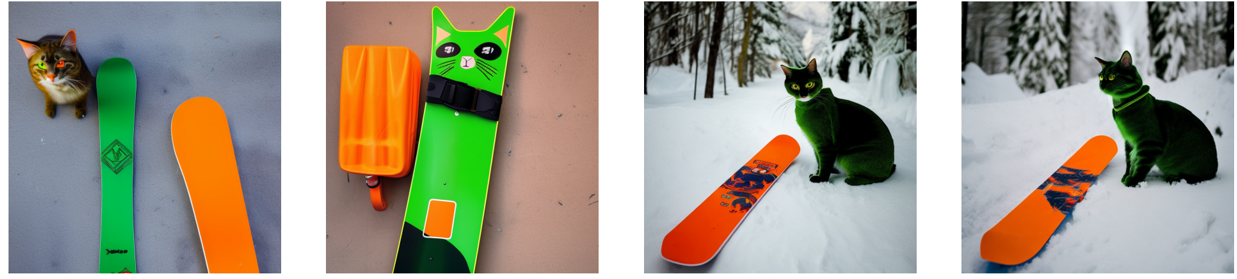}\\
    Prompt: a photo of an orange snowboard and a green cat. Model: SD 2.1
    \includegraphics[width=\textwidth]{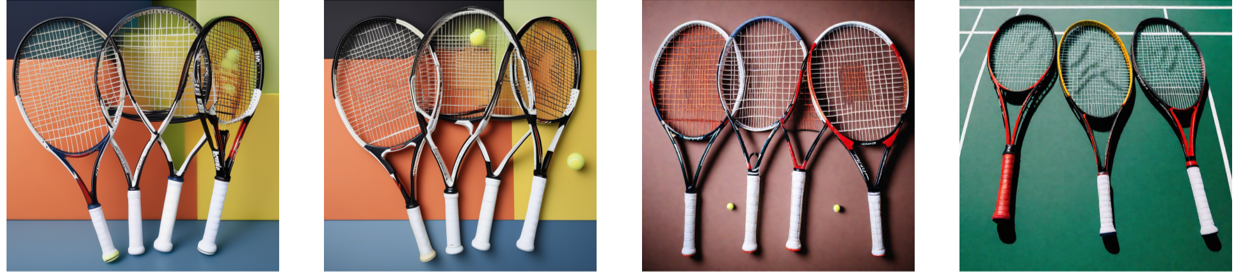}\\
    Prompt: a photo of three tennis rackets. Model: SDXL
\end{figure*}

The first scenario involves generating images that differ substantially from those produced by baseline methods, as shown in Figure~\ref{fig:method_comparison_global}. This outcome is enabled by our method’s ability to expand the initial seed pool via the funnel approach, thereby facilitating the discovery of higher-quality images and enabling identification of a better global mode.

\begin{figure*}[h]
    \centering
    \caption{Local Refinement: Comparison of Proposed Method with Baseline Methods}
    \label{fig:method_comparison_local}
    Best-of-N \hspace{2.5cm}  SMC(M) \hspace{2.5cm} F-SMC(M) \hspace{2.5cm} F-SMC-A(D)\\
    \includegraphics[width=\textwidth]{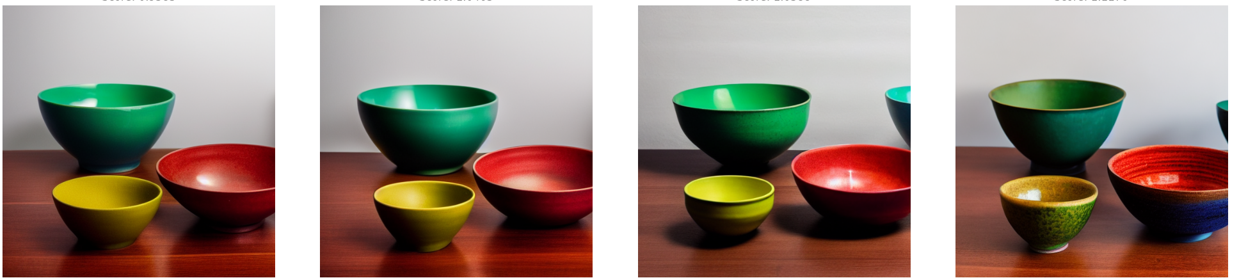}\\
    Prompt: a photo of four bowls. Model: SD1.5\\
    \includegraphics[width=\textwidth]{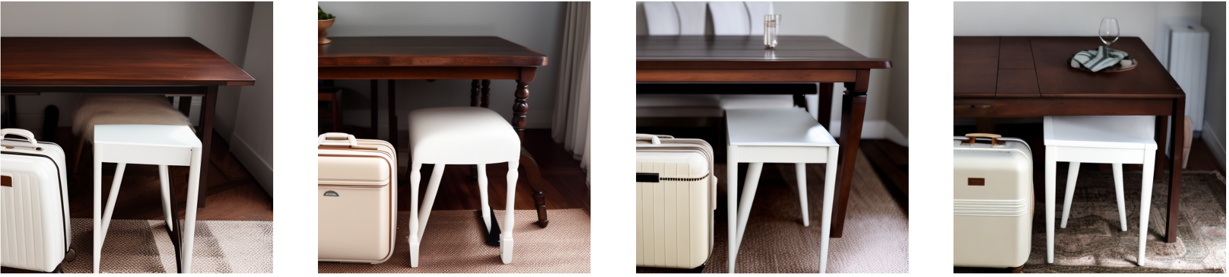}\\
    Prompt: a photo of a brown dining table and a white suitcase. Model: SD2.0\\
    \includegraphics[width=\textwidth]{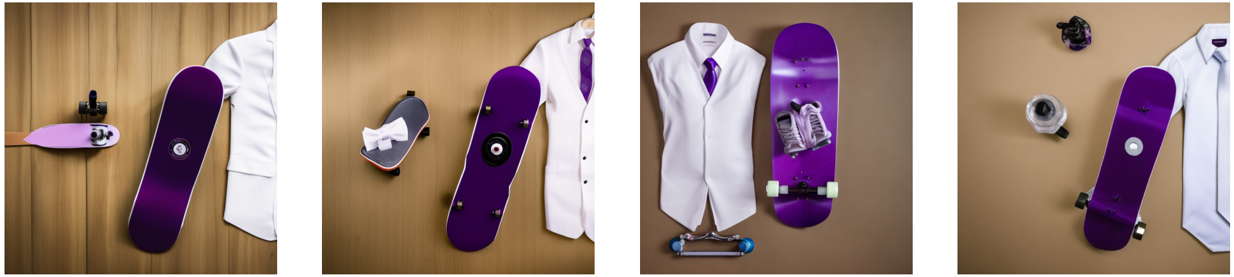}\\
    Prompt: a photo of a white tie and a purple skateboard. Model: SD 2.1
    \includegraphics[width=\textwidth]{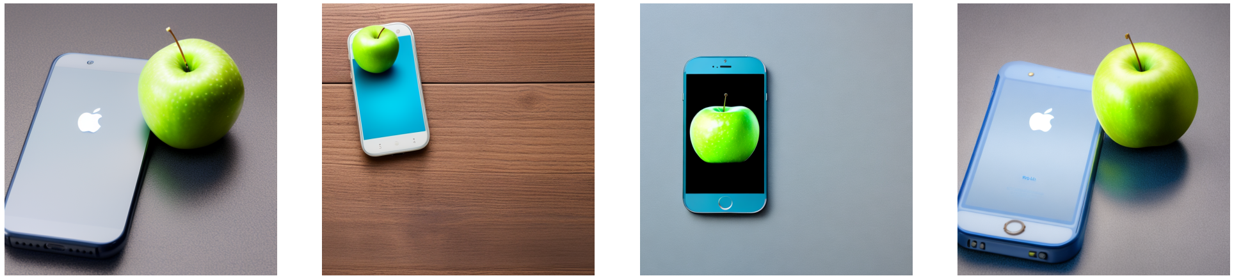}\\
    Prompt: a photo of a blue cell phone and a green apple. Model: SD 2.1
    \includegraphics[width=\textwidth]{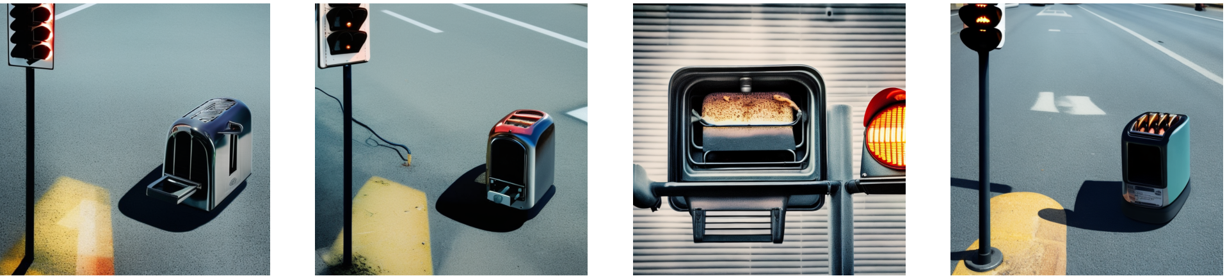}\\
    Prompt: a photo of a toaster below a traffic light. Model: SD 2.1
\end{figure*}

The second scenario, illustrated in Figure~\ref{fig:method_comparison_local}, features cases in which the best image found by our method is visually similar to that produced by baseline methods, indicating a shared underlying noise seed. In such cases, our method outperforms by performing a more fine-grained local search conditioned on the same seed, ultimately yielding a superior image.

\clearpage

\end{document}